\documentclass[manuscript,nonacm]{acmart_modified}

\definecolor{caner}{rgb}{0.36, 0.54, 0.66}
\definecolor{priya}{RGB}{219, 48, 122}
\definecolor{levent}{RGB}{120, 200, 90}

\def\eg{\textit{e.g}\onedot}

\def\etal{\textit{et al}\onedot}

\def \ImNet {ImageNet\xspace}
\def \IG {Instagram\xspace}
\def \CC {Casual Conversations\xspace}
\def \MIAP {OpenImages MIAP\xspace}

\def \harmful {\texttt{harmful}\xspace}
\def \nharmful {\texttt{non-harmful}\xspace}
\def \human {\texttt{human}\xspace}
\def \nhuman {\texttt{non-human}\xspace}
\def \phuman {\texttt{possibly-human}\xspace}
\def \pnhuman {\texttt{possibly-non-human}\xspace}
\def \crime {\texttt{crime}\xspace}
\def \male {\texttt{male}\xspace}
\def \female {\texttt{female}\xspace}

\newcommand{\femaledarker}{\texttt{female darker}\xspace}
\newcommand{\maledarker}{\texttt{male darker}\xspace}
\newcommand{\femalelighter}{\texttt{female lighter}\xspace}
\newcommand{\malelighter}{\texttt{male lighter}\xspace}

\def \oldercc {\texttt{70+}\xspace}

\def \resnet {\texttt{ResNet-50}\xspace}
\def \regnetsmall {\texttt{RegNetY-16}\xspace}
\def \regnetbig {\texttt{RegNetY-128}\xspace}

\newcommand{\ssl}{SSL\xspace}
\newcommand{\wsl}{WSL\xspace}
\newcommand{\fsl}{Supervised\xspace}
\newcommand{\seer}{SEER\xspace}
\newcommand{\swav}{SwAV\xspace}
\newcommand{\supervised}{Supervised\xspace}
\newcommand{\uru}{WSL\xspace}

\newcommand{\africa}{\texttt{Africa}\xspace}

\newcommand{\europe}{\texttt{Europe}\xspace}

\RequirePackage{bbm}
\RequirePackage{xspace}
\RequirePackage{enumitem}

\usepackage{multirow}

\AtBeginDocument{  \providecommand\BibTeX{{    \normalfont B\kern-0.5em{\scshape i\kern-0.25em b}\kern-0.8em\TeX}}}

\makeatletter
\DeclareRobustCommand\onedot{\futurelet\@let@token\@onedot}
\def\@onedot{\ifx\@let@token.\else.\null\fi\xspace}
\makeatother

\usepackage{amsmath}
\usepackage{makecell}
\usepackage{multirow}
\usepackage{colortbl}
\usepackage{capt-of}\usepackage{tabularx}
\usepackage[export]{adjustbox}
\usepackage{subcaption}

\arrayrulecolor{lightgray}

\newlength\savewidth

\newlength\thinwidth

\begin{document}

\title{Fairness Indicators for Systematic Assessments of Visual Feature Extractors}

\author{Priya Goyal}
\email{prigoyal@fb.com}
\author{Adriana Romero-Soriano}
\email{adrianars@fb.com}
\author{Caner Hazirbas}
\email{hazirbas@fb.com}
\author{Levent Sagun}
\email{leventsagun@fb.com}
\author{Nicolas Usunier}
\email{usunier@fb.com}
\affiliation{  \institution{Meta AI Research}
}

\renewcommand{\shortauthors}{Goyal et al.}

\begin{abstract}

Does everyone equally benefit from computer vision systems? Answers to this question become more and more important as computer vision systems are deployed at large scale, and can spark major concerns when they exhibit vast performance discrepancies between people from various demographic and social backgrounds.

Systematic diagnosis of fairness, harms, and biases of computer vision systems is an important step towards building socially responsible systems. To initiate an effort towards standardized fairness audits, we propose \textit{three fairness indicators}, which aim at \textit{quantifying} harms and biases of visual systems. Our indicators use existing publicly available datasets collected for fairness evaluations, 
and focus on three main types of harms and bias identified in the literature, namely \textit{harmful label associations}, \textit{disparity in learned representations of social and demographic traits}, and \textit{biased performance on geographically diverse images from across the world}. We define precise experimental protocols applicable to a wide range of computer vision models. These indicators are part of an ever-evolving suite of fairness probes and are not intended to be a substitute for a thorough analysis of the broader impact of the new computer vision technologies. Yet, we believe it is a necessary first step towards (1) facilitating the widespread adoption and mandate of the fairness assessments in computer vision research, and (2) tracking progress towards building socially responsible models.

To study the practical effectiveness and broad applicability of our proposed indicators to any visual system, we apply them to ``off-the-shelf'' models built using widely adopted model training paradigms which vary in their ability to whether they can predict labels on a given image or only produce the embeddings. We also systematically study the effect of data domain and model size. 
The results of our fairness indicators on these systems suggest that blatant disparities still exist, which highlight the importance on the relationship between the context of the task and contents of a datasets. The code will be released to encourage the use of indicators.

\end{abstract}





\begin{teaserfigure}
  \centering
  \vspace{-3mm}
  \includegraphics[trim= 5 20 5 20, clip, width=0.85\linewidth]{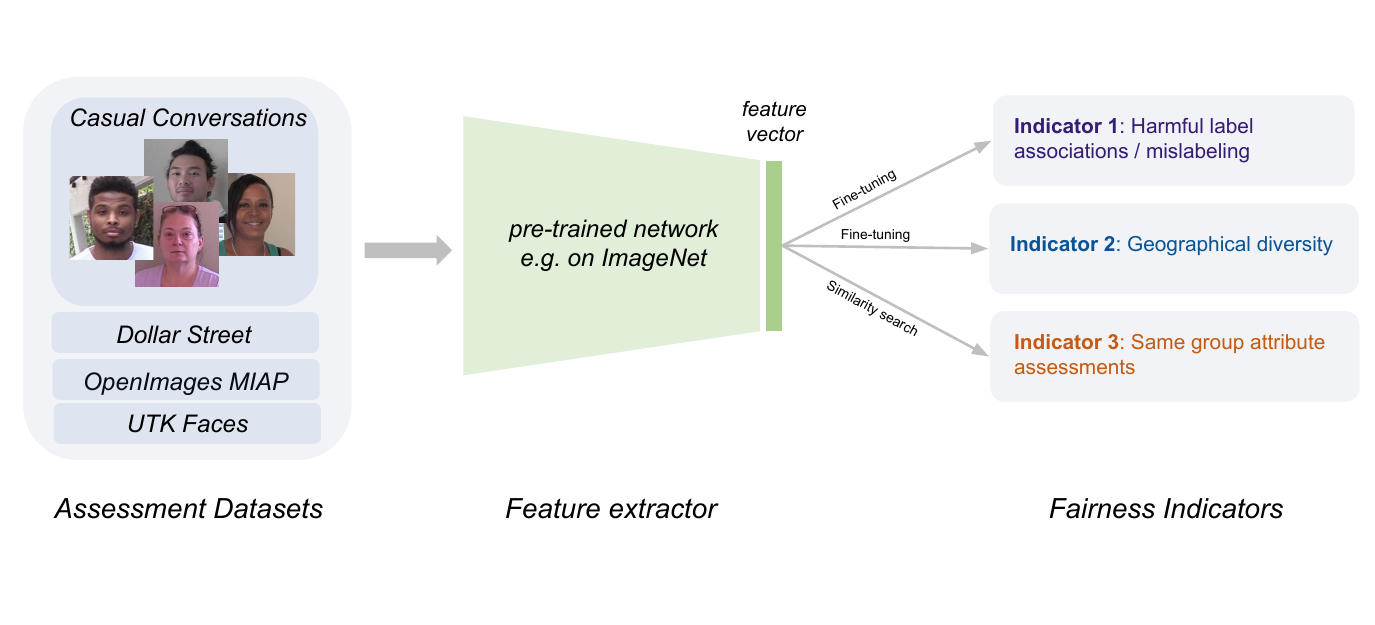}
  \vspace{-3mm}
  \caption{To assess a pre-trained computer vision model for fairness, we take the part of the model that maps given images to its feature space. On the identified \textit{assessment datasets}, we obtain features for images using the given feature extractor. Based on the extracted features, we evaluate fairness through three types of indicators that enable two types of analysis (see Sec.~\ref{sec:indicators}).
    } 
  \label{fig:pull_figure}
\end{teaserfigure}

\maketitle

\newcommand{\nico}[1]{{\color{blue}#1 }}
\newcommand{\nicodone}[1]{{\color{blue} \sout{#1}}}
\newcommand{\ars}[1]{{\color{orange}{#1}}}
\newcommand{\ls}[1]{{\color{green}{#1}}}
\newcommand{\pg}[1]{{\color{cyan}{#1}}}

\section{Introduction}

During the last decade, computer vision systems have been rapidly deployed at large scale in many social contexts, which raised the question of their social impact. One of the main questions is whether these new systems could help resolve social injustice or on the contrary, automate and exacerbate systemic inequality and discrimination \cite{keyes2018, buolamwini2018gender,dentongebru2020,raji20saving}. The study of the bias, or the (un-)fairness, of computer vision systems, has crystallized mostly in the form of black-box audits. These audits typically focus on sensitive groups defined by demographic attributes of people represented in the images or videos, and aim at uncovering discrepancies of error patterns between these groups. As a concrete example, facial recognition applications have been under intense scrutiny because of their questionable usage in surveillance-related applications ~\cite{bookcrawford2021,zuboff2020}, their impact on already marginalized groups \citep{buolamwini2018gender} such as immigrants \cite{harwellwashingtonpostfacial}, among other concerns. More generally, there has been rising concerns regarding image classifiers, even on seemingly mundane tasks such as image tagging~\cite{dentongebru2020}, where studies found significant discrepancies in the error rates in gender classification systems as well as inherent issues in such tasks itself \cite{keyes2018, buolamwini2018gender,hamidi2018gender}, or poor performance when object recognition models are tested on geographically diverse images~\cite{de2019does}.

Increasingly, the developments in computer vision rely on pre-trained feature extractors which are neural networks carefully trained to generate high-level feature representation of images from large images datasets. These feature extractors are then used as the ``backbone'' of classifiers fine-tuned to solve a particular downstream task.  
On tasks such as image classification on ImageNet~\cite{olga2015imagenet}, COCO~\cite{lin2014coco} or few-shot learning, the accuracy of these features extractors has consistently pushed state-of-the-art over the past years  \cite{kolesnikov2020big,mahajan2018exploring,radford2021learning}. 
In conjunction, recent efforts in democratizing computer vision technology were made by open-sourcing feature extractors pre-trained on large datasets. However, recent work showed that the established accuracy measures are far from being reliable indicators for fairness \cite{raji2021ai}. While open-sourcing increases accessibility to models that are hard to train for many ~\citep{mahajan2018exploring,radford2021learning}, the exact context of use of these feature extractors is difficult to anticipate, which makes it even more crucial to understand their potential fairness risks. In particular, we need techniques to thoroughly evaluate the biases and out-of-domain behaviour of these models.

\noindent \textbf{Summary of our contributions.} 
In this paper, we address the problem of assessing the (un-)fairness of feature extractors. To this end, we propose three \textit{fairness indicators}, which aim at quantifying specific harms and biases for certain image based computer vision feature extractors. Our proposed fairness indicators use publicly available datasets previously collected to measure fairness in computer vision \citep{zhifei2017cvpr,dollarstreet,hazirbas2021towards,schumann2021step}, and focus on \textit{systematically} evaluating three main sources of harms that have been identified in the literature:
\begin{enumerate}
    \item \textit{harmful label associations}, where images of people are mistakenly assigned a label that is offensive, derogatory or leads to stereotypes,
    \item \textit{disparity in performance on images from across the world}, following previous studies which showed poor performance on images from outside North-America and Europe, or from low-income households \citep{shankar2017no,de2019does},
    \item  \textit{disparity in learned representations of social and demographic traits} in the pre-trained features, following the analysis of gender-bias in facial recognition systems of \citet{buolamwini2018gender}.
\end{enumerate}

We propose experimental protocols that apply to \textit{any} feature extractor (Figure~\ref{fig:pull_figure}) for which we also provide code and guidance. 
To illustrate how our indicators can be used, we probe fairness of conventional supervised systems trained on ImageNet and two feature extractors trained on millions of internet images using weakly-supervised \citep{mahajan2018exploring} and self-supervised learning \citep{goyal2021self}. 
Our results suggest that compared to supervised training on ImageNet, self-supervised learning on real unfiltered internet data produces significantly fewer errors and smaller discrepancies between sensitive groups. We believe our results will help measure progress towards building fairer models and help facilitate the mandate of fairness audits in further computer vision development.

\noindent \textbf{Limitations and scope of this work.}  On one hand, standardized fairness assessments are appealing to help quantify progress and allow for comparison between models in a reproducible way. Potentially, they could facilitate the widespread adoption of fairness assessments by researchers as the AI research community moves towards inclusive studies with broader impact considerations.
On the other hand, a risk of any benchmark is that it may be confused with an operational definition of fairness, where the sole target is to optimize the few metrics of the benchmark. We emphasize that our fairness indicators are \emph{not} meant to serve as a rigid and comprehensive evaluation of all aspects of fairness. \textit{First}, our limited list of fairness probes cannot capture the multifaceted and ever-evolving aspects of social impacts that computer vision systems can have. \textit{Second}, our indicators are intrinsically limited by the datasets that are currently available. These datasets have limited size, which will make our indicators unreliable if they become the target of optimization. \textit{Third}, the datasets come with limited annotations and follow sampling procedures and definitions of sensitive categories that can themselves be challenged. \looseness-1

We view the concept of standardized fairness probes as an effort towards evaluating harms and tracking progress towards socially responsible models. These fairness probes should consistently evolve as new fairness datasets become available, new concerns regarding the social impact of models are surfaced, new types of model training paradigms are developed or indicators become unreliable or useless. The goal of the benchmark is to facilitate auditing of biases in computer vision models, but it should not be considered as a substitute to the study of broader impact. \\

\noindent \textbf{Overview of the paper.} We discuss the related work in the next section. We then describe the details of our fairness indicators in Sec.~\ref{sec:indicators}. The experimental protocol and results are  presented in Sec.~\ref{sec:experiment_setup} and Sec.~\ref{sec:results} respectively.

\section{Related work}\label{sec:related_work}

\noindent\textbf{Training paradigms in Computer Vision.}
Deep learning along with convolutional architectures and datasets such as ImageNet have shaped modern computer vision~\cite{krizhevsky2012imagenet,he2016resnet,yang2020towards}. Supervised learning has been used as a de-facto approach for training models~\cite{donahue2013decaf,razavian2014cnn}. Several works have demonstrated the benefits of pre-training on large scale curated datasets with weak-supervised learning~\cite{joulin2016learning,mahajan2018exploring,radford2021learning,jia2021scaling}, semi-supervised training~\cite{yan2019clusterfit}, or supervised training on hundreds of millions of images filtered images~\cite{sun2017revisiting,kolesnikov2020big}. Recently, self-supervised learning has been used to train billion-parameter models on billions of internet images~\cite{goyal2021self,caron2020unsupervised}.
A major advantage of large scale models is that the learned visual features can be tuned to work well on a variety of downstream
tasks ~\cite{goyal2021self,mahajan2018exploring,radford2021learning,jia2021scaling}. 

\noindent\textbf{Fairness concerns in Computer Vision.} Although computer vision systems have recently yielded astonishing results,
several societal issues have come to the surface 
as their use materializes in areas like face recognition, self-driving cars and other commercial applications~\cite{book,bookcrawford2021,zuboff2020,eubanks2018,hanna2020}. 

\textbf{i. Harmful label associations.} Recent efforts such as \texttt{ImageNetRoulette}~\citep{roulette2021} have revealed mislabeling of peoples images by computer vision systems when these systems are trained using problematic training data. The incorrect classification and mislabeling of peoples images causes harm and this harm is even greater when the incorrect label corresponds to a \textit{stereotypical or derogatory association}. 

\textbf{ii. Disparity in performances on images from across the world.} Besides mis-classification, image recognition algorithm have also been proven to \textit{not} work equally well across all regions of the world \citep{shankar2017no,de2019does}. This disparity across region of the world has been attributed in part to the datasets and training pipelines based or evaluated on ImageNet, which features mostly images from Western countries \citep{shankar2017no}. 

\textbf{iii. Disparity in learned representations of social and demographic traits of people.} There are growing concerns related to mis-classification of people’s membership in social groups (e.g., gender)~\cite{pinar2021,keyes2018} and the ways that computer vision systems reinforce harmful stereotypes~\cite{doi:10.1177/2378023120967171,bhargava2019exposing}. Raji and Buolamwini~\etal~\cite{raji19actionable} studied impact of Gender Shades~\cite{buolamwini2018gender} in commercial facial analysis and concluded that auditing such systems with the right metrics could potentially reduce the error for marginalized groups,~\eg darker-skinned women. Buolamwini and Gebru~\cite{buolamwini2018gender}, Raji ~\etal~\cite{raji20saving} have shown in their auditing of facial recognition systems that AI systems can discriminate when it comes to gender and race and in particular, found that darker-skinned women are more frequently misgendered and/or not recognized by visual systems.

Motivated by these concerns, we develop two types of indicators as outlined in Figure~\ref{fig:pull_figure}: (Type I) classifier-based indicators built on top of the features and (Type II) similarity-based indicators examining the feature space itself. These two types of indicators cover three types of harmful concerns (discussed in detail in Sec.~\ref{sec:indicators}).

\noindent\textbf{The impact of broader impact statements.} Broader impact statement requirement and more recently, ethics review processes have brought a much needed perspective to machine learning research community and its effects are spreading to wider groups of researchers \cite{prunkl2021institutionalizing,ethicsReview}. A particular example is the recent Open AI CLIP~\cite{radford2021learning} model which is a large scale model pre-trained on wide variety of images with language supervision. In its broader impact section, the authors present fairness evaluations of their model on \textit{harmful label associations} and \textit{disparity in gender recognition} using FairFace~\cite{karkkainen2021fairface} dataset. However, these evaluations did not provide systematic protocols that can be followed for any pretrained model for assessing fairness such as geodiversity. The evaluations are also based on prompt engineering where the input is "text" as prompt instead of the visual features which is a unique property of the model and does not generally apply to computer vision systems. Further, the assessment on gender recognition disparity involved training / predicting gender which has ethical concerns \citep{raji20saving}. Compared to this work, we are interested in protocols that allow researchers and practitioners to audit and compare fairness of any CV system on several types of harms/biases and \textit{without} requiring training an attribute (age, gender, skintone etc.) classifier.

\begin{table}[t]
    \centering
    \footnotesize{
    \setlength{\tabcolsep}{0.9em}\scalebox{0.9}{
        \begin{tabular}{ll}
        \toprule
        \textbf{Dataset} & \textbf{Description} \\
        
        \toprule
        \texttt{Casual Conversations} & \makecell[l]{\textasciitilde3K images containing face crops of people's faces. 
         \textit{self-identified} gender  (`male', `female', \\ `other' and `n/a'), 
         age (from $18$ - $85$) and annotated Fitzpatrick skin tone labels. \\  Only used for model inference and \textit{not} for training} \\

        \midrule
        \texttt{OpenImages MIAP} & \makecell[l]{ \textasciitilde44K bounding boxes of peoples images from the \texttt{test} set. 
         \textit{perceived} gender (predominantly masculine, \\ predominantly feminine, unknown), 
         \textit{perceived} age range (young, middle, older, unknown) \\ Only for inference on the bounding boxes with height and width >= 100 in the test set }\\

        \midrule
        \texttt{UTK Faces} & \makecell[l]{\textasciitilde24K face images with \textit{apparent} age, race and gender. \\ Only apparent gender labels are used since data is not balanced wrt skin-tone }\\
        
        \midrule
        \texttt{Dollar Street} & \makecell[l]{\textasciitilde16K images, 108 concepts, $54$ countries,
         $4$ regions (The Americas, Africas, Asia, Europe), \\
         $289$ households with different income levels, on average 53 unique images per household.) \\ 108 concepts are mapped to 94 classes in ImageNet} \\
        
        \bottomrule
    \end{tabular}
    }
    \vspace{-0.01in}
    \caption{List of \textbf{Fairness datasets} used in the proposed indicators in Sec.~\ref{sec:indicators}.}
    \label{tab:fairness_datasets}
    \vspace{-9.6mm}
        }
\end{table}

\noindent\textbf{Datasets for measuring fairness in computer vision.} 
Several fairness evaluation datasets have been proposed to facilitate fairness assessment by enabling testing of classification performance on images from diverse geographic locations~\cite{shankar2017no} or correlation between detection performance and an \textit{income} variable of the object~\cite{de2019does}. Recent work emphasized the importance of how \textit{people} images are classified or otherwise analyzed by computer vision systems from early datasets of faces with geographically diverse collection~\cite{klare2015pushing,levi2015age} or \citet{buolamwini2018gender}'s intersectional benchmark to the recent datasets FairFace~\cite{karkkainen2021fairface}, Casual Conversations~\cite{hazirbas2021towards} and More Inclusive Images for People (MIAP)~\cite{schumann2021step}. These works offer curated datasets with labels obtained through clear annotation rules and with specific efforts deployed for checking annotation bias. 

We describe the datasets we use in Table~\ref{tab:fairness_datasets}: Casual Conversations~\cite{hazirbas2021towards}, OpenImages MIAP~\cite{schumann2021step}, and UTK Faces~\cite{zhifei2017cvpr} contain images of people and are used in the indicators of harmful label association and/or the same-group similarity search. DollarStreet \citep{dollarstreet,de2019does}, is used in the geographical fairness indicator. A breakdown of number of samples per attribute can be found in Table~\ref{tab:datasets_distribution} and Table~\ref{tab:dollar_street_dataset_distribution} and detailed descriptions of the datasets can be found in the Appendix \ref{sec:detailed_datasets}. \\

\noindent We further discuss other fairness studies pertaining to criticism of ImageNet, fairness metrics and centering fairness around the context of the task in Appendix~\ref{sec:appendix_further_related_work}.

\section{Fairness Indicators}
\label{sec:indicators}

\begin{table}[t]
    \centering
    \resizebox{\linewidth}{!}{
    \begin{tabular}{l*{5}c}
        \toprule
        \textbf{Indicator} & \textbf{Dataset(s)}& \textbf{Task Type} & \textbf{Goal} & \textbf{Sensitive group(s)} & \textbf{Metric}\\
        
        \toprule
        \texttt{\textbf{Label Association}}
        & \makecell{\textit{Casual Conversations }\\\textit{OpenImages MIAP}
        } & \makecell{image\\ classification}
        & \makecell{Measure association between \texttt{harmful} \\ predictions and sensitive groups of people}
        & \makecell{gender, skin tone and age (CC)\\ gender and age (MIAP)}
        & \makecell{\% \texttt{harmful} predictions\\ at various confidence thresholds. }
        \\
        \midrule
        \texttt{\textbf{Geo Diversity}} & \makecell{\textit{Dollar Street}}
        & \makecell{image\\ classification} & \makecell{Measure disparities in object recognition \\depending on household income}
        & \makecell{income households\\ and region of the world}
        & \makecell{hit rate (object recognition)} \\
        
        \midrule
        \texttt{\textbf{Same attribute}} & \makecell{Database:\textit{UTK-Faces} \\ \textit{Queries:Casual Conversations}}
        & \makecell{similarity\\ search} & \makecell{Measure disparities between sensitive groups \\ in learned representations of images of people}
        & \makecell{gender, skin tone, age}
        & \makecell{\texttt{Precision@K}}
        \\
        \bottomrule
        
    \end{tabular}}
    \vspace{-0.01in}
    \caption{A summary of the fairness indicators proposed in Sec.~\ref{sec:indicators}.}
    \label{tab:fairness_indicators_summary}
    \vspace{-8.mm}
    \end{table}
Following the three main sources of harms and discrepancies between groups outlined in Sec.~\ref{sec:related_work}, we propose three fairness indicators that apply to pre-trained feature extractors. 
\begin{itemize}
    \item The first two indicators (Indicator 1 and 2 in Fig.~\ref{fig:pull_figure}) perform an indirect evaluation of feature extractors using \textit{classifiers} built by fine-tuning the feature extractors. We propose two measurements that assess out-of-domain generalization of the classifiers:
        \begin{enumerate}
            \item \textit{harmful mislabeling} of images of people (Sec.~\ref{sec:label_association}),
            \item \textit{geographical disparity} in object recognition (Sec.~\ref{sec:describe_dollar_street_indicator}).
        \end{enumerate}
    \item The third indicator (Indicator 3 in Fig.~\ref{fig:pull_figure}) performs a direct evaluation of the extracted features using a similarity search task. It aims at measuring \textit{disparities in learned visual representations of social memberships of people} (Sec.~\ref{sec:similarity_search}). 
\end{itemize}

\noindent A summary of the high-level design of the indicators is given in Table \ref{tab:fairness_indicators_summary}. The details of our indicators are discussed below, together with the main differences from the variants that have been proposed in the literature. We also note the limitations of the indicator and intended use in Appendix~\ref{sec:appendix_limitations}.

\subsection{Indicator1: Harmful label association}
\label{sec:label_association}

The goal of the harmful label association indicator is to study how much classification algorithms make potentially harmful and biased label associations on images of people for various subgroups (age, gender, skintone). We describe all the components of the indicator: the datasets, the definition of harmful associations, the sensitive groups and the metrics.\looseness-1
    
\begin{itemize}
    \item \textbf{Requirement.} This indicator requires a visual system that has label prediction capability. We discuss in Appendix~\ref{sec:appendix_finetune_system} how one can adapt certain systems (such as those trained with self-supervision) to predict labels if the system doesn't have this capability. 
    \item \textbf{Datasets.} We design two independent tests using two different datasets: 
        \begin{itemize}
            \item \textbf{Casual Conversations}: which contains \textit{faces} of people,
            \item \textbf{OpenImages MIAP}: which contain more diverse images that represent close-to real-world scenarios. 
        \end{itemize}
    We \textit{emphasize} that these datasets are used for \textit{inference only} and the classifiers should not be trained (or pre-trained) on these datasets, as the indicator is designed to stress-test classifiers on out-of-domain images.
    
    \begin{table}[!t]
    \centering
    {\small
    \setlength\tabcolsep{8pt}
    \begin{tabular}{ll}
        \toprule
        \textbf{Association type} & \textbf{Labels in the ImageNet taxonomy} \\
        \toprule
        \texttt{Human} & face, people\\
        \midrule
        \texttt{Possibly Human} & makeup, khimar, beard \\
        \midrule
        \texttt{Non-Human} & \makecell[l]{swine, slug, snake, monkey, lemur, chimpanzee, \\
        baboon, animal, bonobo, mandrill, rat, dog (CC only), \\
        capuchin, gorilla, mountain gorilla, ape, great ape, orangutan.} \\
        \midrule
        \texttt{Possibly Non-Human} & dog, cat \\
        \midrule
        \texttt{Crime} & prison \\
        \bottomrule
    \end{tabular}
    \vspace{-0.01in}
    \caption{Label mapping from ImageNet taxonomy for computing harmful label associations (Sec.~\ref{sec:label_association}). ``Dog'' is considered as a \textit{Non-Human} label on Casual Conversations v1 only, and the \texttt{Possibly Non-Human} type is evaluated solely on OpenImages MIAP. We manually mapped a subset of labels in full ImageNet taxonomy~\cite{imagenetTaxonomy} to these association types.}
    \label{tab:label_association}
    \vspace{-9.mm}
    }
\end{table}
    \item \textbf{Label Associations.} After performing inference with the classifier on any of the datasets (OpenImages MIAP or CC), we measure harmful label associations in a specific label taxonomy by first categorizing the labels into three main \textit{types} of interest: \texttt{Human}, \texttt{Non-Human} and \texttt{Crime} in order to study derogatory stereotypes. The predictions that fall into the \texttt{Non-Human} and \texttt{Crime} categories are considered harmful associations. The exact mapping between labels and the association types depend on the underlying taxonomy. The meaning of different association types is:
            \begin{itemize}[leftmargin=*]
            \item \texttt{Human} are labels in ``people'' subtree. In our test, predicting a \texttt{Human} label should be considered correct.
            \item \texttt{Non-Human} labels are labels in the ``animal'' subtree that would lead to harm if predicted for certain sensitive groups.\looseness-1
            \item \texttt{Crime} labels are \textit{not} in ``people'' nor ``animal'' subtrees but, if predicted, would lead to harm/bias.
        \end{itemize}
    Labels that do not fall into the three types above are neither counted as correct or incorrect. Yet, in order to better account for possible mistakes or correct predictions, we also record the following labels:
        \begin{itemize}[leftmargin=*]
            \item \texttt{Possibly-Human} labels are \textit{not} in ``people'' subtree but describe ``people'' attributes (\textit{e.g.,}  ``makeup'' or ``beard'')
            \item \texttt{Possibly-Non-Human} labels are non-human labels that may be legitimately predicted. For instance, it is common in OpenImages MIAP that a person is holding a pet. In that case, \texttt{cat} might be considered a correct prediction. This type does not apply to the CC dataset which only contains faces of people.
        \end{itemize}
    Assessing the correctness of either \texttt{Possibly-Human} or \texttt{Possibly-Non-Human} predictions requires manual visual inspection. 
        The exact mapping between labels and the label types depend on the underlying taxonomy. We provide the label types for the ImageNet taxonomy used in our experiments in Table~\ref{tab:label_association}.
  
    \item \textbf{Sensitive groups.} The harmful predictions are measured on images of people belonging to various sensitive groups (which depends on the information available in the dataset). We use two datasets: Casual Conversations (CC) and OpenImages MIAP. For each dataset, sensitive groups and data distribution is detailed in Table~\ref{tab:datasets_distribution}.\looseness-1
                                    
    \item \textbf{Metrics - confidence rated predictions}. For all subgroups, we consider the top-$5$ predicted labels by the classifier and report the percentage of images of that subgroup for which at least one predicted label falls into each label type at a certain confidence threshold of label prediction.

    While top-$5$ prediction is a common metric in research on image classification~\cite{he2016resnet}, classifiers also have prediction probabilities for each label, which can be used as a confidence score.
    
    Assessment of classifiers with varying confidence score thresholds follows the literature on \textit{selective classification} \citep{el2010foundations,geifman2017selective}, also called classification with a reject option \citep{herbei2006classification} or classification with abstention \citep{zhang2016extended,schreuder2021classification}. We argue for using the \textit{confidence-based assessment}: 
        \begin{itemize}
            \item the evaluations \textit{without} considering confidence scores, do not distinguish between mistakes that the classifier is very confident in, compared to mistakes where the model has very low confidence (which can be treated automatically for instance by sending to a human annotator, or simply not considering the image for further evaluation)
            \item it does affect the harms/biases conclusions regarding the fairness of models (if a model predicts a harmful label but with a very low confidence score (say 0.02), accepting low confidence predictions increases harm).
            \item we believe this is particularly relevant to out-of-domain tests (and, similarly, in deployed systems that may receive out-of-domain data) since confidence scores are also used to detect out-of-domain samples \citep{lee2018simple}.
        \end{itemize}
    \textbf{Choosing thresholds.} using confidence scores introduces an additional burden of choosing the threshold - a problem that is often referred to as the \textit{risk-coverage trade-off}: higher threshold leads to less mis-classification, but also less coverage because the classifiers abstains from making predictions on more images. Since a classifier that constantly abstains is useless, we need to choose a non-trivial threshold in practice. The choice of the threshold is inherently problem/task dependent (depends on the potential risks of mis-classification, including but not exclusively fairness/harms risks). Providing general guidance on how to solve this trade-off in context is out of the scope of this paper, and hence we report results for different thresholds.
    
    \begin{table}[t]
    \centering
    \small{
    \setlength{\tabcolsep}{0.3em}\scalebox{0.8}{
    \begin{tabular}{l  cc c cccc c cc}
        
         & \multicolumn{2}{c}{\textbf{Gender}}  && \multicolumn{4}{c}{\textbf{Age}} && \multicolumn{2}{c}{\textbf{Skin Tone}} \\
        
        \cmidrule{2-3} \cmidrule{5-8} \cmidrule{10-11}

         \textbf{Dataset} & \textbf{feminine labels} & \textbf{masculine labels} && \textbf{18-30} & \textbf{30-45} & \textbf{45-70} & \textbf{70+} && \textbf{lighter} & \textbf{darker} \\
        
        \midrule
        \texttt{CC} &  $1,627$ &  $1,294$ && 931 & 1046 & 870 & 62  && 1646 & 1329 \\
        
                \texttt{UTK Faces} & $11,525$ &  $12,583$ && 7728 & 5727 & 4712 & 1414 && n/a & n/a \\

        \midrule
         &  &  && \textbf{young} & \textbf{middle} & \textbf{older} & \textbf{unknown} &&  &  \\

         \midrule
        \texttt{OpenImages MIAP} & $10,807$ & $14,345$ && $3,754$ & $23,966$ & $986$ & $14,817$ && n/a & n/a \\

        \bottomrule
    \end{tabular}
    }
    \caption{Number of samples of each characteristic in the datasets. Gender labels are self-identified-\{female, male\} for CC, \{predominantly feminine, predominantly masculine\} for OpenImages, and apparent-\{female, male\} for UTK. On CC, we follow~\citet{buolamwini2018gender} and group the six-point Fitzpatrick scale into two types: \texttt{Lighter} (Type I to Type III) and \texttt{Darker} (Type IV to Type VI) and group age into four groups \texttt{18-30, 30-45, 45-70, 70+}. On the OpenImages MIAP, there are three perceived gender subgroups (\texttt{predominantly masculine, predominantly feminine, unknown gender}) and four perceived age subgroups (\texttt{young, middle, older, unknown}).
    }
    \label{tab:datasets_distribution}
    \vspace{-9.6mm}
}
\end{table}

    \item \textbf{Summary.} Overall, using the indicator involves the following steps:
    \begin{itemize}
        \item \textbf{Step1}: For a given taxonomy, generate the label associations. For ImageNet, we provide the list in Table~\ref{tab:label_association}.
        \item \textbf{Step2}: Run the model \textit{inference only} on each image in the datasets and capture the top-$5$ model prediction along with the confidence scores.
        \item \textbf{Step3}: For different subgroups, measure the percentage of images labeled with different association types for different confidence thresholds.
    \end{itemize}
    \item \textbf{Difference with the literature.} Label association tests were already present in the analysis of CLIP \citep{radford2021learning}. The main differences with our proposal are discussed in Sec.~\ref{sec:related_work} under "The impact of broader impact statements". In short, our approach (i) applies to any visual extractor, (ii) aims at comparing different models, (iii) uses self-identified gender in CC dataset and wider variety of practical images present in OpenImages MIAP.
                        On a related note, we also mention that \citet{yang2020towards} proposed a revisited ImageNet by filtering out \textit{unsafe} labels. This is different from \texttt{harmful / biased} associations, which we study in our work. For instance, labels such as \texttt{gorilla} are marked \textit{safe} in \citet{olga2015imagenet} because they are legitimate labels on images of the corresponding animal. However, they are clearly \textit{harmful} when predicted on images of people of certain groups, as is evident in historical incidents~\cite{googlemisclassification}.
\end{itemize}

\begin{table}[t]
    \centering
    \footnotesize{
    \setlength{\tabcolsep}{0.9em}\scalebox{0.8}{
    \begin{tabular}{lc c ccccc}
    
        \toprule
        \multicolumn{2}{c}{\textbf{income}} && \multicolumn{5}{c}{\textbf{region}}\\
        bucket &    range (\$) && \textbf{Africa} &   \textbf{Asia} &  \textbf{Europe} &  \textbf{Americas} & \textbf{Global}\\
        
        \midrule
        \texttt{low}&  $27-90$           &&    37 &   20 & 0 &  5 & 62   \\
        \texttt{medium} & $93-1,700$     &&    23 &  111 & 17 & 26 & 177 \\
        \texttt{high}   & $1,700-10,000$ &&     3 &   17 & 17 & 13 & 50   \\
        
        \midrule
        total & -- && 63 & 148 & 34 & 44 & 289\\
        \bottomrule

    \end{tabular}
    }
    \caption{Number of samples of each characteristic in the datasets. Number of households per region (as defined by \citet{de2019does}) and per income buckets on the Dollar Street dataset.}
    \label{tab:dollar_street_dataset_distribution}   
    \vspace{-4.0mm}
    }
\end{table}
\subsection{Indicator 2: Geographical diversity and fairness}
\label{sec:describe_dollar_street_indicator}
This second indicator aims at assessing the object recognition accuracy of visual systems on images from around the world. Similar to the label association indicator in Sec~\ref{sec:label_association}, this indicator assesses classifiers. We share details of all components of this indicator.

\begin{itemize}
    \item \textbf{Requirement.} This indicator requires a visual system that has label prediction capability. See Appendix~\ref{sec:appendix_finetune_system} for how one can adapt certain systems (such as those trained with self-supervision) to predict labels if the system doesn't have this capability and in particular, predict labels in ~\texttt{Dollar Street} taxonomy.
    
    \item \textbf{Dataset.} We use \texttt{Dollar Street} dataset and mapped the initial annotations to the ImageNet taxonomy as described in Appendix \ref{sec:detailed_datasets}\footnote{It is possible to use other taxonomies by mapping them to the original Dollar Street annotations.}. The images in this dataset are annotated with the label, country, region and the household income (for the household that image represents). 
    
    We \textit{emphasize} that the classifiers should \textit{not} be trained or pre-trained on the \texttt{Dollar Street} dataset, since this indicator aims at stress-testing classifiers on an unseen, diverse set of images.
    
    \item \noindent \textbf{Sensitive groups.} We propose 2 different sensitive sub-groups:
        \begin{itemize}[leftmargin=*]
            \item \textit{Regions of World}: \texttt{The Americas, Europe, Asia} and \texttt{Africas}.
            \item \textit{Household income buckets}: Given the household income (in USD), we group the income into buckets as \texttt{round(log(household income)/3)}. Despite the simplicity of this formula, this bucketing yields three income buckets across the full dataset, which allows us to simplify the analysis into: \texttt{low, medium} and \texttt{high income} groups. The distribution of the number of households by region/income buckets is given in Table \ref{tab:dollar_street_dataset_distribution}. The choice of income as a sensitive feature follows \citet{de2019does} which showed that usual classification models perform significantly better on images from high-income thresholds. 
                    \end{itemize}
    
        \item \textbf{Metrics.} In the ~\texttt{Dollar Street} dataset, relatively few households (in total $289$) are represented, but with a rather high number of images per household. The work by \citet{de2019does} computed the mean income of represented households in each country of the dataset, and counted the average hit rate over images from that country where an image is counted as a \textit{hit} if one of the top-$5$ predictions is the ground truth annotation. 
        
        In our work, for reliable fairness audit, we take an alternative approach to computing average hit rates. Our approach aims at being less sensitive to spurious correlations that are due to same-household. To that end, our metrics use the following two pre-processing steps. 
        \begin{enumerate}
            \item First, we observed that for some households, the same image appears several times with different labels. Since our classifiers are not meant for predicting multiple labels for the same image (ImageNet classifiers are typically trained for single-label prediction), we first de-duplicate images, counting the image as correctly classified if \textit{any} of its ground truth labels has been predicted. This leaves $15,222$ images for $289$ households.
            
            \item Second, since images from the same household tend to be visually much more similar than images from different households, we first compute the hit rate on each household (as the average hit rate over the images of this households), and then take the average of these per-households hit rates over the various sensitive groups (region, income bucket or  income bucket$\times$ region). 
        \end{enumerate}
    
        \item \textbf{Summary.} We summarize the end-to-end process:
            \begin{itemize}
                \item \textbf{Step 1: Adaptation of Visual systems to predict labels in Dollar Street taxonomy.} Follow the details in Appendix~\ref{sec:appendix_finetune_system}.
                \item \textbf{Step 2: Inference on Dollar Street images.} Use the visual system to predict labels and the confidence score of label prediction for each image in the Dollar Street dataset.
            \item \textbf{Step 3: Compute the metrics.} This is computed as described above.
            \end{itemize}
    \item \textbf{Difference with the literature.} 
        The difference with \citet{de2019does} lies in the metrics (computed over households) and our definitions of sensitive groups (regions and income buckets). We demonstrate our definitions allow for more reliable performance estimates (discounting the correlation between images introduced by the household) and confidence intervals (individual countries, as used by \citet{de2019does} do not contain enough households to produce meaningful confidence intervals).
\end{itemize}

\subsection{Indicator 3: Same-attribute assessment via similarity search}\label{sec:similarity_search}

The goal of this indicator is to probe the visual systems for disparity in learned representations of images of people based on an attribute (age, gender, skintone, race etc.). We follow an instance retrieval approach which involves a \texttt{Database} and \texttt{Queries}. The \texttt{Database} is where we search and the \texttt{Queries} are the inputs for which we want to retrieve similar things (for example embedding vectors) from the \texttt{Database}. This indicator evaluates a pre-trained feature extractor by performing similarity search given the queries in the \texttt{Database}, where the similarity is defined as the \textit{cosine similarity} of images in embedding space. We describe the components of this indicator in detail: 

\begin{itemize}
    \item \textbf{Requirement.} Any visual system that needs to be audited. Unlike previous indicator, the system does \textit{not} need to have a classifier.
    \item \textbf{Datasets.} We use UTK Faces as the \texttt{Database} and Casual Conversations (CCv1) as the \texttt{Queries}. Both datasets have almost balanced representation of different genders and age groups, so there is no significant representation bias present in these datasets. 
    
    \item \textbf{Sensitive groups.} For CC, similar to the Indicator 1, we use the sensitive groups described in Table~\ref{tab:datasets_distribution}. 
    
    \item \textbf{Metrics.} Since our task is similarity search, we measure \texttt{Precision@K} which measures proportion of $K$ most similar images that have the same gender as the query image. We focus here on gender since it is the only common attribute between UTK Faces and CC. 
    
    \item \textbf{Summary.} Given a visual feature extractor, the end to end process is as follows:
        \begin{itemize}
            \item \textbf{Step 1: Extract model embeddings.} Run the inference on the images of the UTK Faces and CC datasets and save the model output / embeddings.
            \item \textbf{Step 2: Perform similarity search and measure.} First, normalize the embeddings to unit L$2$-norm. Then, for each image embedding in the CC dataset, perform similarity search using the UTK Faces embeddings and then compute the \texttt{Precision@K} metric for each sensitive subgroup.
        \end{itemize}
    \item \textbf{Difference with the literature.} In the analysis of CLIP, \citet{radford2021learning} propose a gender classification task on FairFace.  Our proposal differs from their in the use of a \texttt{similarity search} task rather than building a classifier of sensitive demographic attributes. Our choice is motivated by two important aspects:
        \begin{itemize}
            \item \textit{first}, training attribute specific classifiers (such as gender, age, skintone etc) is increasingly contrary to intended uses of datasets collected for fairness, \citep{hazirbas2021towards,schumann2021step},
            \item \textit{second}, and relatedly, there is rising concerns regarding training classifiers for sensitive labels such as age, gender, skin tone. We believe that a fairness assessment should avoid relying on building intermediate questionable artifacts (such as a gender recognition system) when possible. The similarity search example corresponds to uses cases such as image retrieval~\citep{berman2019multigrain}, where we would want to account for same-group similarity when the query image contains people.
        \end{itemize}
\end{itemize}

\section{Experimental Setup}
\label{sec:experiment_setup}

\begin{figure}[t!]
    \centering
    \vspace{-3mm}
    \includegraphics[width=0.75\textwidth]{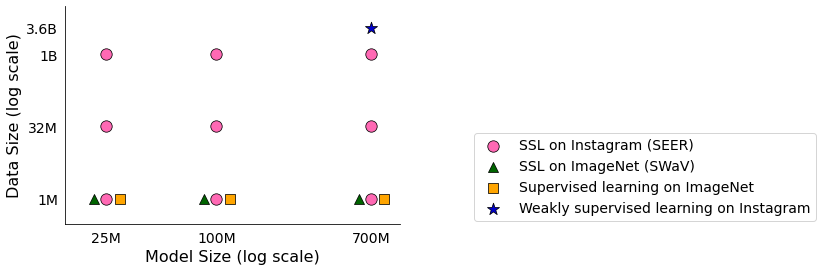}
    \vspace{-3mm}
    \caption{
    Grid of model and data sizes (in log-log scale) of different \textbf{training paradigms} compared in Sec.~\ref{sec:experiment_setup}.
        } 
        \label{fig:model_data_comparisons}
\end{figure}
    
We illustrate the use of the indicators presented in the previous section by comparing three types of feature extractors trained within different paradigms and at different data/model scale and different data domain:
\begin{itemize}
    \item \textbf{Supervised training on ImageNet:} Our baseline feature extractors follow the standard practice of training a neural network classifier on ImageNet \citep{imagenet_cvpr09} and considering the layer before the prediction heads as a feature extractor \citep[see e.g.,][]{huh2016makes}. For the reference, ImageNet contains $1.28$~million images classified with a single label that belongs to a taxonomy of $1,000$ labels, derived from WordNet~\citep{wordnetbradford2020}.
    
    \item \textbf{Weakly-supervised training on filtered Instagram data:} We study an open source feature extractor, \uru \citep{mahajan2018exploring}, that was trained on $3.6$ Billion public Instagram images. The training data is constructed by utilizing the hashtags associated with images and filtering the images with hashtags that have synsets in WordNet~\cite{wordnetbradford2020} resulting in $27,000$ hashtags. The convolutional neural network was trained using supervised learning using the hashtags associated to each image as target labels. Similar to supervised ImageNet models, the feature extractor is the last layer of the network before the prediction head. We take this model as representative of feature extractors trained at large scale with \textit{weak} supervision. 
    
    \item \textbf{Self-Supervised training on ImageNet or uncurated Instagram data:} We study two representative approaches for self-supervised training, \swav \citep{caron2020unsupervised} and \seer \citep{goyal2021self}. They share the same underlying self-supervised training principles, based on constrained clustering of image crops. The main difference between the two is that \swav has hyperparameters optimized for self-training on ImageNet, whereas \seer was tuned for training models on random internet image at a much larger data scale. Similar to \uru, \seer is trained on public Instagram images. However, in contrast to \uru, the dataset used for \seer is a completely random subset of $1$ billion images that underwent \textit{no data filtering or curation} whereas in \uru, images were filtered based on hashtags that match English nouns in WordNet.
\end{itemize}
We take as main representative of these approaches a large convolutional neural network of the RegNet family~\cite{radosavovic2020designing}, \regnetbig which has $700$~Million parameters. We chose this model because this model is available off-the-shelf for various training paradigms (supervised, weakly-supervised and self-supervised) making it possible to compare fairness of different training paradigms. We also study smaller RegNet and \resnet in Appendix~\ref{sec:label_assoc_model_capacity}.

An important part of our experimental is an in-depth ablation study. We believe that it is reasonable to compare weakly/self-supervised training at the billion-image scale to supervised training because weak- and self-supervision are precisely meant to enable such scale in contrast to supervised learning where collecting large amount of curated labeled data is simply infeasible. Further, as data domain and data size change within training paradigms, we propose an in-depth study where we control additional parameters. In this ablation study, we focus on \supervised training and \swav/\seer as we have pretrained vision systems available off-the-shelf allowing us to save compute resources and also \seer seemed to achieve overall better results on fairness indicators than \uru. We study systematically two axes:
\begin{itemize}
    \item \textbf{Data size and domain.} We evaluate \seer models on three subsamples of the Instagram data, containing $1$ million, $32$ million and $1$ billion images to study the effect of scaling the dataset size. Notice that since ImageNet contains about $1$ million examples, comparing \seer trained on \IG with $1$ million examples and \swav essentially compares the effect of the data domain.
\item \textbf{Model size and architectures.} We study model size ranges from $25~Million$ to $700~Million$ parameters. We primarily focus on Convolutional Neural Networks (ConvNets)~\cite{cnn1989} in our experiments.\footnote{Our indicators can readily be used with other models such as Vision Transformers~\cite{dosovitskiy2021image}}. We chose \resnet~\cite{he2016resnet} as an example of lightweight model, and two sizes of RegNet, \regnetsmall ($100$ million parameters) and \regnetbig~\cite{radosavovic2020designing}.
\end{itemize}
We systematically evaluate all combinations of \seer models for all data size and models, and all models for \supervised training and \swav. A pictorial summary of all models studied is given in Fig.~\ref{fig:model_data_comparisons}.

\section{Results and observations}\label{sec:results}
In this section, we present our findings and provide a comprehensive analysis of our experiments.

\begin{figure}[t!]
\centering
\begin{subfigure}{.45\textwidth}
    \centering
    
     \includegraphics[width=1.\textwidth]{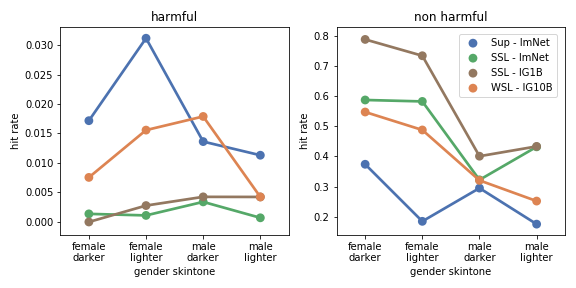} 
     
     \includegraphics[width=1.\textwidth]{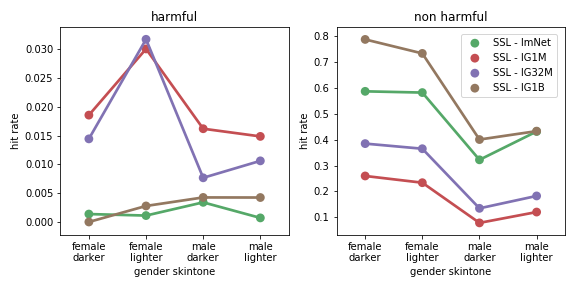}    

    \includegraphics[width=1.\textwidth]{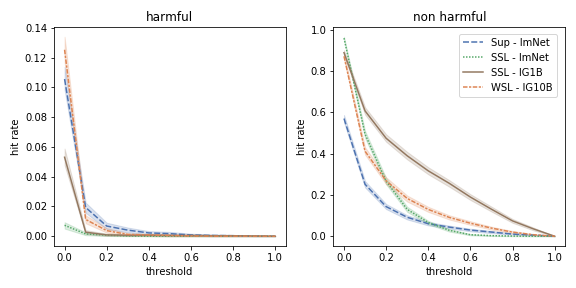} 
    \caption{\CC.}
    \label{fig:cc_label_assoc}
\end{subfigure}
\begin{subfigure}{.45\textwidth}
    \centering
     \includegraphics[width=1.\textwidth]{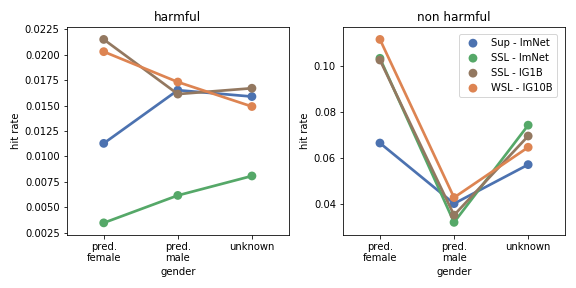}  
     
     \includegraphics[width=1.\textwidth]{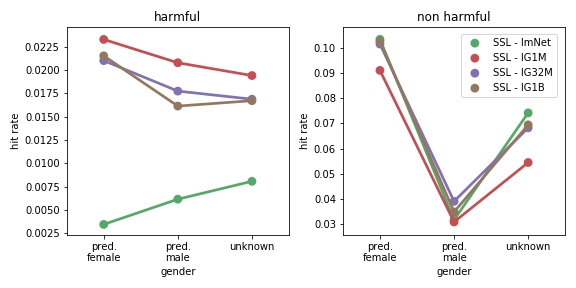}    

    \includegraphics[width=1.\textwidth]{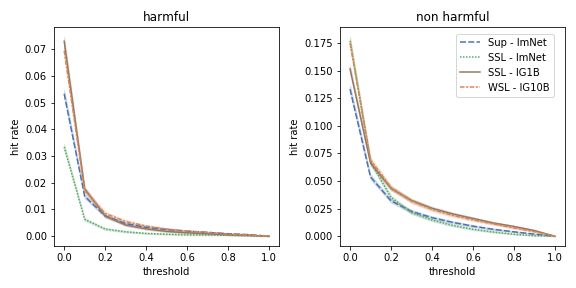} 
    \caption{\MIAP}
    \label{fig:miap_label_assoc}
\end{subfigure}
\caption{Label association results: (Top)  Effect of training paradigm; (Mid) Effect of data size and domain; (Bottom) Confidence of models. For \texttt{harmful} association, lower hit-rate is better. For \texttt{non-harmful} association, higher hit-rate is better. }
\label{fig:label_assoc}.
\end{figure}

\subsection{Label association results}
\label{ssec:labelassoc}

This section analyses the impact of training paradigm, data size and domain on the potentially harmful label associations. Figure~\ref{fig:label_assoc} depicts results on \CC and \MIAP. Results are reported for the \regnetbig model and are stratified into groups -- based on gender and skintone on \CC and on perceived gender in \MIAP. Additional results stratified into age groups can be found in the Appendix~\ref{sec:label_assoc_age_group}. We consider \harmful and \nharmful predictions with a minimum confidence threshold $\tau = 0.1$. Note that \harmful labels include \crime and \nhuman, whereas \nharmful labels only include \human.

\textbf{Effect of training paradigm.}
Figures~\ref{fig:cc_label_assoc}--\ref{fig:miap_label_assoc} (top row) compare the effect of different training paradigms on \CC and \MIAP, respectively. We consider models pretrained on \ImNet and on the larger scale \IG $1$B. 
On \textit{\CC}, trends appear to favor the \ssl paradigm, which results in the lowest \harmful and the highest \nharmful hit rates. Leveraging supervised signals (\fsl, \wsl) results in more harmful associations and larger \harmful association hit rate differences across different groups. We also note that \ssl-\ImNet leads to overall slightly lower \harmful predictions than \ssl-\IG (possibly because \ImNet data domain is more object centric), but the trend is reverted for \nharmful predictions where the \ssl-\IG obtains notably higher hit rates (possibly because \IG data domain is more human-centric). 
In \textit{\MIAP}, \ssl-\ImNet continues to yield to lowest \harmful hit rates. However, in this case, \fsl-\ImNet appears to follow. Interestingly, models pretrained on \IG tend to be among the top \harmful label predictors, no matter the training paradigm, suggesting that in the case of \MIAP the data may play a more important role than the training paradigm. For \nharmful predictions, \ssl models tend to yield higher hit rates, which are also comparable to those of \wsl. It is worth noting that the effect of training paradigm may strongly depend on the model capacity. In particular, we note that when using a smaller \resnet, \fsl consistently becomes more competitive on both datasets considered, lowering its \harmful hit rates, and reaching among the highest \nharmful hit rates across all groups. See Appendix~\ref{sec:label_assoc_model_capacity} for a detailed analysis on the effect of model capacity.

\textbf{Effect of data size and domain.} Since \ssl learning allows training models on unbounded data size, we study this effect for \ssl models specifically. Figures~\ref{fig:cc_label_assoc}--\ref{fig:miap_label_assoc} (mid row) show how data domain and data size affect the training of \ssl models on \CC and \MIAP, respectively. Both datasets exhibit similar trends. When comparing \ssl models trained on different \IG data sizes, it appears that models benefit from additional data: using $1$B data points increases the \nharmful predictions and lowers the \harmful ones, especially when compared to the model trained on $1$M data points. These gains appear more evident on \CC than on \MIAP, where improvements are somewhat modest. However, in both datasets, the \ssl-\ImNet results in comparable or lower \harmful prediction hit rates than the best \IG-based model, suggesting that within self-supervised training paradigm, the data content might be more critical than the scale of the data to mitigate potentially harmful associations.\looseness-1

\textbf{Effect of prediction confidence of models.} We assess how different training paradigms impact the confidence of models in Figures~\ref{fig:cc_label_assoc}--\ref{fig:miap_label_assoc} (bottom row), for CC and \MIAP respectively. bu studying how of \harmful and \nharmful label predictions change as a function of increasing the confidence threshold $\tau$ that a prediction is considered ``correct''. On \textit{CC}, we observe that \ssl models consistently require a small confidence threshold to push \harmful predictions rates close to $0$. Moreover, both \fsl and \wsl models start off with notably higher \harmful prediction hit rates and take longer to lower their prediction rate close to $0$, with the \fsl model being slightly more confident than the \wsl model overall. For \nharmful predictions, \ssl-\IG model consistently results in more confident predictions than any other model for $\tau > 0.1$. In this case, \ssl-\ImNet appears to be less confident than \wsl-\IG, and reaches \nharmful hit rates which are close to $0$ faster. \fsl consistently yields the lowest \nharmful hit rates. On \textit{\MIAP}, only the \ssl- \ImNet consistently presents the lowest \harmful hit rates, and requires a rather small threshold value to bring the hit rate very close to $0$. \fsl-\ImNet starts with lower \harmful hit rates that \ssl-\IG, which are however quickly matched by both (\ssl and \wsl) \IG based models, suggesting that \fsl is overall more confident than \IG based models. Models pretrained on \IG all exhibit a similar behavior for \harmful hit rate on \MIAP, with \wsl being slightly more confident in \harmful predictions than \wsl. For \nharmful predictions, similar to CC, \ssl-\IG is the most confident model, resulting in high hit rates as we increase the threshold, which are only matched by \wsl pretrained on \IG. Finally, the confidence of models also depends on their capacity. 
See Appendix~\ref{sec:label_assoc_model_capacity} for a detailed analysis on the effect of model capacity.\looseness-1

\begin{table*}[t]
\begin{tabularx}{\textwidth}{*{2}{>{\centering\arraybackslash}X}}
  \centering
    \includegraphics[width=\linewidth,valign=B]{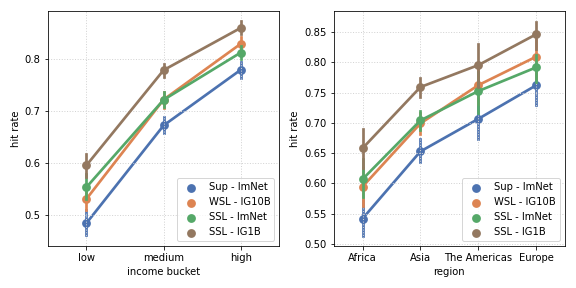}
    \captionof{figure}{Geodiversity, hit rates for \supervised, \wsl and \ssl.
    \label{fig:geodiversity:best}}
&
    \includegraphics[width=\linewidth,valign=B]{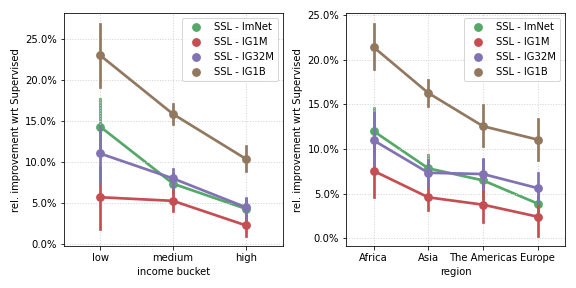}
  \captionof{figure}{Geodiversity, relative improvement with respect to \supervised on \ImNet. Effect of data domain and size on \ssl. \label{fig:geodiversity:modelanddata}}
\end{tabularx}
\end{table*}

\subsection{Geographical Fairness Results}

In this section we analyze the results on the geodiversity indicator. The main results are summarized in Fig.~\ref{fig:geodiversity:best} and Fig.~\ref{fig:geodiversity:modelanddata}. All results in this section use \regnetbig. We refer the reader to App.~\ref{sec:geodiversity_model_capacity} for a study of the effect of model capacity.

\noindent \textbf{Effect of training paradigm.} Fig.~\ref{fig:geodiversity:best} plots the hit rate for \supervised on \ImNet, \wsl, and \ssl on \ImNet and \IG. We observe that there are drastic discrepancies in performances across sensitive groups, with a difference of $0.25$ between lower and higher income budgets, and about $0.20$ between \africa and \europe. Nonetheless, all versions of \wsl and \ssl significantly improve over \supervised training, over all sensitive groups (income buckets and region). Moreover, \ssl trained on \IG is significantly better than the other models on all groups. 

\noindent \textbf{Effect of data domain and scale.} Fig.~\ref{fig:geodiversity:modelanddata} plots the \textit{relative improvement in hit rate} with respect to \supervised training on \ImNet, for four \ssl models: one trained on \ImNet, and three models trained on \IG with $1$M, $32$M and $1$B examples respectively. We observe that increasing the data size on \IG has a positive effect on performances, with a large gap between $32$M and $1$B examples. \ssl on \ImNet seems better than \ssl on \IG with $1$M examples, which suggests \ImNet is not a worse data domain than \IG for this task. Yet, the \textit{first} conclusion is that these result support the idea that training on large, diverse datasets improves generalization on geographically diverse images. 
The \textit{second} important conclusion is that both \ssl models have \textit{higher} relative improvement on \textit{lower} income buckets, and generally higher relative improvement on groups with worse performances overall. Even though there are still large discrepancies between senstive groups -- across the world and by income buckets, \ssl seems a promising avenue to reduce discrepancies in performances between groups.

\subsection{Results of Same Attribute Retrieval using Similarity Search}\label{sec:simsearch}

We now present the results for the last indicator, where the embeddings are directly used in a similarity search algorithm.

The main results are summarized in Fig.~\ref{fig:simseach:main}, which shows the precision-at-10 and precision-at-50 broken down by gender, skin tone and gender$\times$ skintone image groups. As previously, we compare four \regnetbig models, \supervised on \ImNet, \wsl, \ssl on \ImNet and \ssl on \IG. 
For all models, the precision-at-10 and the precision-at-50 exhibit very similar trends, so we discuss both at the same time under the generic name ``precision''. Detailed results depending on model capacity can be found in Appendix~\ref{sec:simsearch_model_capacity}.

\noindent \textbf{Effect of training paradigm.} For all models, there are discrepancies in the retrieval of same-gender images between query images of females and males, as well as between query images of people with darker and lighter skin tone. Looking at the breakdown by gender$\times$ skin tone, we see that models work roughly the same on \maledarker and \malelighter, and the precision drops for female darker. The models have different discrepancy profiles. Both \supervised and \ssl trained on \ImNet exhibit large discrepancies in precision between \male and \female queries ($0.25$ and $0.20$ respectively). \wsl reduces this discrepancy to $0.11$, and overall significantly reduces discrepancies in precision across all gender$\times$skin tone groups. \ssl trained on \IG drastically reduces discrepancies while improving the precision, reaching with \femalelighter, \maledarker  and \malelighter reaching $~95\%$ precision. The precision for \femaledarker queries still lags $10$ points behind, which shows that there are still blatant differences to address.

\noindent \textbf{Effect of data domain and scale.} In Fig.~\ref{fig:simseach:datasize}, we show the precision-at-10 for \regnetbig trained with \ssl on \ImNet and three different data sizes for \IG data ($1$M, $32$M and $1$B images). The precision at other rank thresholds yields similar trends. Training on \ImNet and \IG with $1$M examples leads to similar precision values, which suggests that the difference in data domain between \ImNet and \IG has little impact on this indicator. We observe a large increase in precision when the data size increases to $32$M, with little differences between $32$M and $1$B. These results suggest that large datasets are critical in improving the fairness indicator. This conclusion is in-line with the geodiversity indicator, even though there is no improvement when increasing the data size from $32$M to $1$B.

\begin{table*}[t]
\begin{tabularx}{\textwidth}{>{\arraybackslash\centering\hsize=0.66\hsize}X>{\arraybackslash\centering\hsize=0.33\hsize}X} \centering
   \includegraphics[width=0.62\textwidth]{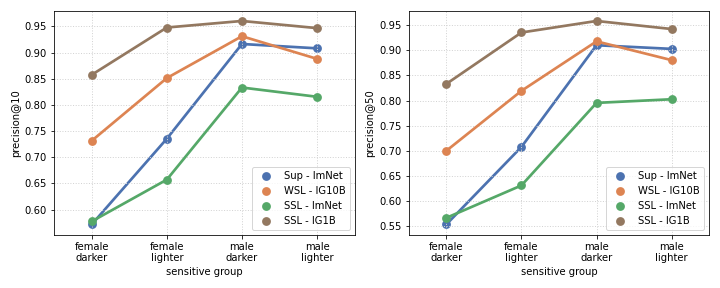}
    \captionof{figure}{Same-group similarity search, precision@10 (left) and precision@50 (right) for the four main models compared.\label{fig:simseach:main}}
&
\centering
\includegraphics[width=0.31\textwidth]{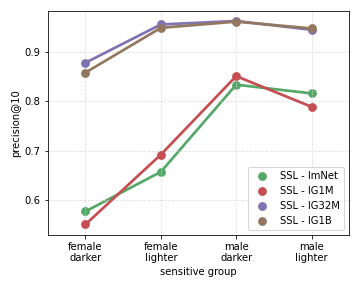}
   \captionof{figure}{Similarity search, varying the data domain and size for \ssl with \regnetbig.\label{fig:simseach:datasize}}
\end{tabularx}
\end{table*}

\subsection{Summary of All Indicator results}
    
       \begin{enumerate}
           \item Self-Supervised training on larger and diverse datasets such as Instagram data leads to most improvement in generalization on geographically diverse images. The performance improves the most for lower-income and non-Western regions of the world. 
           \item SSL paradigm leads to highest \nharmful and lowest \harmful label associations. Instagram data leads to highest \nharmful label associations. Further, data scale increases the \nharmful associations. 
           \item Self-Supervised training on Instagram data leads to drastic reduction in discrepancies in the same gender retrieval while achieving high precision for different gender and skintone which improved further with data scale.
                       \end{enumerate}

\section{Conclusion and Future work}
Measuring fairness of computer vision systems in a systematic way is necessary to build technologies that are fairer, more inclusive for all people from different demographics. In order to spur the progress in systematic assessment of bias in computer vision systems, we propose three \textit{fairness indicators} that are designed to probe main sources of biases in computer vision models. We apply our indicators on most commonly used deep architectures that are trained using different training paradigms on various data sources, and assess the generalization of these models on four publicly available fairness datasets. Our comprehensive analyses show that large models using vast amounts of data (without requiring any annotations or labels) perform best across all subgroups defined in fairness datasets. We hope to spur further research in the field and hence advance model diversity and generalization to people across sensitive groups. Our assessment is intended to be used in conjunction with qualitative analysis of models’ broader impact.

\noindent \textbf{Future work.}
In this work, we only focus on models trained on either uni-labeled \ImNet (one label per image) or unlabeled \IG data, and evaluate the fairness on the datasets that provide either fairness labels,~\eg age, gender, skin tone, or concepts,~\eg income with geo-location. One open question for future research is to carry further analyses on the pre-trained models that are trained on multi-labeled datasets (multiple label per image).

\begin{acks}
The authors thank Adina Williams, Armand Joulin, Laurens van der Maaten, Melissa Hall, Miranda Bogen, Piotr Bojanowski, David Lopez-Paz and Quentin Duval for their feedback on earlier versions of the paper.
\end{acks}

\bibliographystyle{ACM-Reference-Format}
\bibliography{facct_strip}

\appendix

\section{Detailed descriptions of datasets}
\label{sec:detailed_datasets}

\noindent\textbf{Casual Conversations (CC)}~\cite{hazirbas2021towards} consists of $45,186$ videos of paid participants across five different cities in the US. Dataset provides \textit{self-provided} age and gender labels in addition to annotated Fitzpatrick skin tone~\cite{FitzPatrick75} and ambient lighting conditions. \textit{Age} category varies from $18$ to $85$,~\textit{gender} is limited to `male', `female', `other' and `n/a' (not available, prefer not to say) and there is only a few videos for `other' and `n/a' gender categories. Nevertheless, CC is one of the few public datasets that has \textit{self-identified} age and gender labels and hence we decided to assess our indicators also on this dataset. For the purpose of our indicators, we employ the CC dataset for model inference only and \textit{do not} perform model training on this dataset. See Appendix~\ref{sec:ccdetails} for more details how to retrieve face crops. \\

\noindent\textbf{OpenImages MIAP}~\cite{schumann2021step} is a dataset designed to enable ML Fairness and constructed by providing additional annotations for a subset of OpenImages v6~\cite{openimagesv4} dataset. These annotations include bounding boxes of people and attribute labels for fairness such as \textit{perceived} gender (predominantly masculine, predominantly feminine, unknown) presentation and \textit{perceived} age range (young, middle, older, unknown).

For the purpose of our indicators, we use OpenImages MIAP \textit{test} set for inference only. The \textit{test} split contains $22,590$ images where each image has multiple bounded boxes. In order to focus on images of single people, we perform inference only on the bounding boxes with height and width $>=100$ as otherwise the people are barely visible. We also filter out bounding boxes with gender or age of the \texttt{unknown} category as we observed they often are sketches, images of objects, or images where only the lower body is visible. The use of these images not critical in our fairness study, and we preferred keeping them out since it is unclear what label to expect from the classifier. Overall, we perform inference on $43,523$ effectively cropped images. We share more details on how we carefully apply the bounding box crop for inference in Appendix~\ref{sec:miapdetails}. \\

\noindent\textbf{UTKFaces}~\cite{zhifei2017cvpr} is a large-scale face dataset that consists of $24,108$ face images each of which is annotated for \textit{apparent} age, race and gender. The term  \textit{apparent} denotes that the labels are not  \textit{self-identified}. The dataset has almost balanced data between for male / female apparent genders with the downside of no data for non-binary gender. Images cover a variety of variations in pose, facial expression, illumination, occlusion and resolution. For the purpose of our proposed fairness indicator, we only use the apparent gender labels for which the dataset is also balanced. \\

\noindent\textbf{Dollar Street}~\cite{dollarstreet} is a collection of $16,073$ images of households captured all around the world by a group of photographers. The images capture 108 concepts from $289$ households in $54$ different countries across the world. The countries represent 4 different regions (The Americas, Europe, Asia, Africa). We manually mapped the 108 concepts to 94 classes in ImageNet~\cite{olga2015imagenet} dataset~\footnote{We release the full data information including the label mapping file to enable use of this indicator.}. The households have income levels (measured in USD) varying between $67\$$ and $10k$. There are on average $~53$ unique images per household, with a maximum of $135$ labeled objects per household.

\section{Experimental details}

\subsection{Casual Conversations v1}\label{sec:ccdetails}

For our experiments, we use the \textit{mini} version of the dataset which consists of $2,982$ videos (two videos per participant with one~\textit{dark} and one~\textit{bright} lighting video when possible). Note that face crops are not available for $29$ videos where DLIB~\cite{King2009} has failed to detect any faces. For each video, following~\cite{hazirbas2021towards}, bounding boxes are detected on each frame of the video using DLIB~\cite{King2009} that are resized by a factor of $0.5$ for faster computation and then upscaled by $2$. In order to increase the area of face and background (hair and clothes) we enlarge the bounding boxes by a factor of $1.5$ from the center on the aligned frames and resize the face crops to $256x256$. We take the middle frame of the video where there exists a face crop (when possible) and perform inference resulting on $2,982$ face images. 

\subsection{OpenImages}\label{sec:miapdetails}

For OpenImages, we filter out bounding boxes with height or width smaller than $100$ pixels, because we observed they correspond to barely visible people. We also filter out bounding boxes with gender or age of the \texttt{unknown} category as we observed they often are sketches, images of objects, or images where only the lower body is visible. The use of these images not critical in our fairness study, and we preferred keeping them out since it is unclear what label to expect from the classifier.

We then followed a two-step process to have bounding boxes fit the $224\times 224$-input of our models:
\begin{enumerate}
    \item if the box's shape is too far from a square (we use the rule $\frac{\max(\text{height, width})}{\min(\text{height, width})} \geq 1.2$), we crop the bounding box further to a square from the top-left of the bounding box.
    
    That way, we obtain a square input with focus on the top of the bounding box, which usually is the head for images of people. 
    
    \item we then resize the resulting image to 224x224
\end{enumerate}

\subsection{Adapting a visual system to predict labels}\label{sec:appendix_finetune_system}
For the indicators proposed in Sec~\ref{sec:label_association} and Sec.~\ref{sec:describe_dollar_street_indicator}, visual systems need to be able to predict the labels and in particular labels in Dollar Street taxonomy for Geographical fairness indicator. There are 2 possible scenarios:
\begin{itemize}
    \item \textbf{Visual system has label prediction capability (for Dollar Street taxonomy)}: If the dataset taxonomy of the dataset that the visual system is trained on, already predicts (Dollar Street) labels, such system can be used directly for inference on (Dollar Street) test images in \textit{zero-shot} manner.
    \item \textbf{Visual systems don't have label prediction capability (and/or don't predict Dollar Street taxonomy)}: For some visual systems (such as self-supervised models) which are not inherently trained to predict any labels, the models can be adapted (for example by finetuning) to predict the labels by training on a \textit{subset} of datasets like ImageNet~\cite{olga2015imagenet}~\footnote{We release the information of exact subset of ~20K subset images from ImageNet which capture the Dollar Street taxonomy and also subset labels that correspond to label associations in ~\ref{sec:label_association} indicator.}.
\end{itemize}
 We also acknowledge that the caveat of using ImageNet as the transfer dataset can introduce the potential bias in the system but we note that this strategy can still enable comparisons of several visual systems conditioned that all the visual systems are adapted to Dollar Street taxonomy using the same training process (irrespective of if the models already have label prediction capability). We further note that our 3rd indicator allows to measure fairness purely from the raw model embeddings.

\section{Limitations in more detail}
\label{sec:appendix_limitations}
We note several limitations of our proposed indicators below, and some of them are the inevitable result of using currently available datasets. 
We note that these limitations would be easily overcome with more diverse and fairer datasets, which consist of all possible inclusive labels for all attributes.
\begin{itemize}
    \item The proposed indicators provide a proof-of-concept for what could be the systematic assessment and evaluation of visual systems by utilizing the existing fairness datasets. These evaluations can enable comparison of  models and measure how well they are calibrated (in particular, how do they impact marginalized populations). However, given a visual system, the choice of what indicator to measure depends on the context and this choice must be thoroughly assessed with proper stakeholder involvement so as to answer why those indicators are chosen, what kind of assumptions are embedded in this choice, and what specific questions do the system designers aim to answer \cite{paml, kalluri2020don}. 
    
    \item We further note that these fairness indicators are complementary to the model and dataset documentation. These indicators \textbf{\textit{do not}} replace proper documentation of dataset building and model reporting practices, but it can come together as a standard way of outlining a baseline comparison when one is developing or deploying new models. Furthermore, the proposed indicators can evolve/expand through time to include new types of visual systems, new fairness probes and new datasets as deemed appropriate.
    
    \item In our analysis of several visual systems, we only considered one model in each setup and didn't evaluate fairness if different seeds are used to initialize the models as this is beyond the scope of this work but we encourage using multiple seeds when probing the model fairness with the indicators.
    
    \item we focus on single-label prediction, mostly for convenience because classification is the most studied CV task in the literature. Even focusing only on classification, single-label prediction makes label ambiguity a problem -- without all labels that could be considered correct for an image, error rates are unreliable.
    
    \item some limitations from the use of dataset:
        \begin{itemize}
            \item the definitions of attributes differ across datasets which limits the possibility of cross-dataset tests (for instance, UTK faces doesn't have Fitzpatrick skintone scale in contrast to Casual Conversations. This limits the study to gender and age attributes in our Similarity search based indicator.
            \item most dataset like UTK-Faces don't yet have data for non-binary genders which strictly limits the fairness probe for all social memberships.
            \item most datasets only provide \textit{perceived} or \textit{apparent} labels for people's social membership which itself could be biased.
            \item lack of labels social membership labels (age, gender, skintone etc) puts constraints on the type of task we can address. For instance, publicly available Hate Speech datasets~\cite{kiela2020hateful,kiela2021hateful} do not have labels for sensitive groups, and hence we can't analyze hate speech harms/biases of current visual systems towards specific groups this data.
        \end{itemize}
\end{itemize}

\section{Further context and related work}
\label{sec:appendix_further_related_work}
\noindent\textbf{Criticism of ImageNet.} ImageNet~\cite{imagenet_cvpr09,olga2015imagenet} has spurred immense developments and advances in Computer Vision over decades. However, recently Yang~\etal~\cite{yang2020towards} pointed out several reasons that ImageNet~\cite{imagenet_cvpr09} might cause potential bias and therefore harm in the downstream models. Dulhanty
and Wong~\cite{dulhanty2019auditing} studied the demographics of people in ImageNet dataset by using computer vision models to predict the gender and age of people, and demonstrated that, e.g., males aged
15 to 29 make up the largest subgroup. Stock and Cisse~\cite{stock2018convnets} did not explicitly analyze the dataset but demonstrate that models
trained on ImageNet exhibit misclassifications consistent with racial
stereotypes. Steed~\etal~\cite{steed2021image} further showed that unsupervised models trained on ImageNet can automatically learn racial, gender and intersectional biases from the way people portrayed in images that were curated from the web. Recently, effort has been made by Yang~\etal~\cite{yang2020towards} to reduce the bias in ImageNet dataset that resulted in removal of 2,702 \texttt{synsets} (out of 2,800 total) from \texttt{person subtree} in ImageNet dataset. \

\noindent\textbf{Fairness metrics for machine learning models.} Most of the recent works utilize average performance indicators broken down by sensitive groups. The metrics themselves measure disparities between groups in terms of predicted positive value (\texttt{PPV}), True/False positive rates (\texttt{TPR}, \texttt{FPR})~\cite{buolamwini2018gender}, error rates~\citep{de2019does} or risk difference~\cite{mehrotra2021mitigating}. There is no consensus nor general guidance for choosing the metric depending on context. Despite incompatibilities between metrics~\cite{kleinberg2018inherent,chouldechova2017fair} or otherwise counterarguments regarding the use of these aggregate performance measures~\cite{corbett2018measure,golz2019paradoxes}, there has been little debate on which metric to use for fairness audits in computer vision. This contrasts with the analysis of the COMPASS risk scores, which spurred an intense debate over metrics~\cite{ProPublicaAnalysis,dieterich2016compas,flores2016false} (even though this debate is for the most part unresolved). \

\noindent\textbf{Centering fairness around the context of the task.} Proper documentation of models and datasets are key priorities in developing new AI systems and should be carried out concurrently to benchmarks. 
Comprehensive approaches range from Model cards~\cite{modelcardmichell2019} aiming to standardize transparent model reporting, and data documentation~\citep{hutchinson2021accountability} encouraging accountability at every stage of the data collection process. Raji~\etal~\cite{raji20saving} pointed out five \textit{ethical concerns} that should be taken into consideration while developing products for algorithmic auditing in order to prevent harms on protected groups. A common aspect of these developments is rooted in making the \textit{context} of the task and the underlying cultural / social context as an important factor in developing transparent and accountable machine learning systems. 
Without such considerations, relying only on abstract notions of fairness measures fails to address the core problem of the developed system~\cite{mitchell50years2019}. These challenges impact the researcher and the practitioner alike and the standardized protocols enabling systematic fairness assessment is a long overdue step forward for computer vision developments.\looseness-1

\section{Additional label association results}
\label{sec:appendix_label_assoc_result}

In this section we provide additional label association results on \CC and \MIAP. 

\subsection{Results for different age groups.}
\label{sec:label_assoc_age_group}
Following Section~\ref{sec:label_association}, all results are reported for the \regnetbig model, considering predictions with a minimum confidence threshold $\tau = 0.1$.\looseness-1

\begin{figure}[t!]
\centering
\begin{subfigure}{.45\textwidth}
    \centering
    
     \includegraphics[width=1.\textwidth]{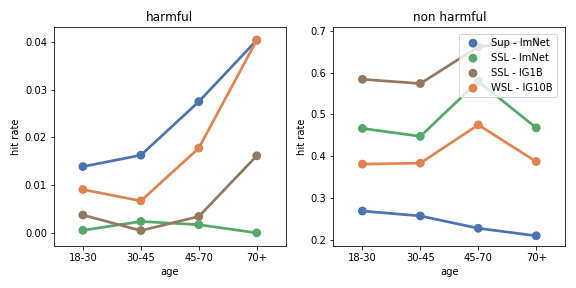} 
     
     \includegraphics[width=1.\textwidth]{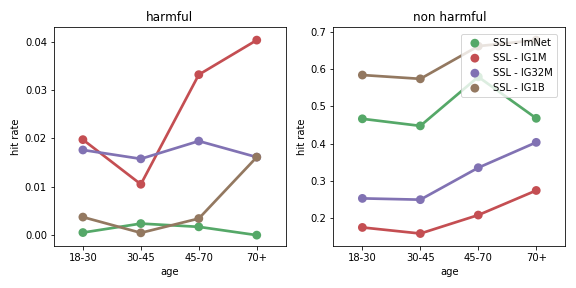}    

    \caption{\CC.}
    \label{fig:age_cc_label_assoc}
\end{subfigure}
\begin{subfigure}{.45\textwidth}
    \centering
     \includegraphics[width=1.\textwidth]{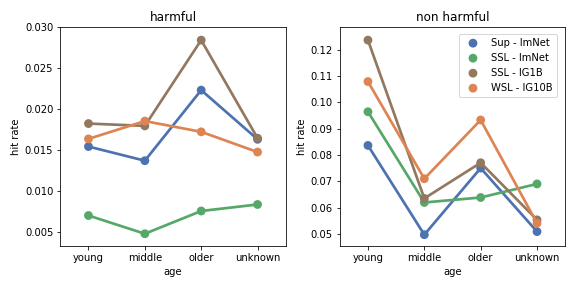}  
     
     \includegraphics[width=1.\textwidth]{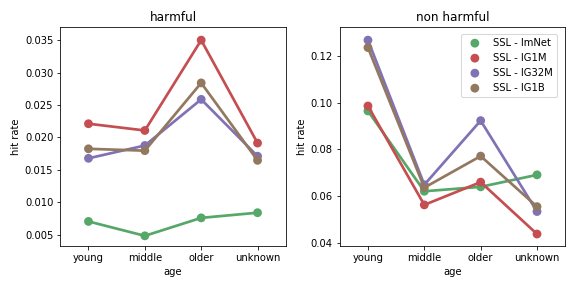}    

    \caption{\MIAP}
    \label{fig:age_miap_label_assoc}
\end{subfigure}
\caption{Label association results stratified into groups based on age: (Top)  Effect of training paradigm; (Bottom) Effect of data size and domain. For \harmful labels, the lower the hit rate the better; conversely, for \nharmful labels the higher the hit rate the better.\looseness-1 }
\label{fig:age_label_assoc}
\end{figure}

Figure~\ref{fig:age_label_assoc} depicts the impact of training paradigm, data size and domain on the potentially harmful label associations on both \CC and \MIAP when stratifying results based on (perceived) age. Results include \harmful (\crime, \nhuman) and \nharmful (\human) predictions.\looseness-1

\textbf{Effect of training paradigm.}
Figures~\ref{fig:age_cc_label_assoc}--\ref{fig:age_miap_label_assoc} (top) compare the effect of different training paradigms on \CC and \MIAP, respectively when stratifying results based on different age groups. As in the gender-skin tone stratification, we observe that trends appear to favor the \ssl paradigm on \CC. However, we note that in this case, \ssl-\ImNet reaches similar \harmful hit rates as \ssl-\IG, except for the \oldercc age group, where the \harmful predictions are lower. The trend observed in the \nharmful predictions is the same as for the gender-skin tone stratification, with \ssl-\IG obtaining notably higher hit rates than \ssl-\ImNet (possibly because \IG data domain is more human-centric). In \textit{\MIAP}, the analysis of the results also yields to similar observations as in the gender-skin tone classification. In particular, \ssl-\ImNet continues to yield the lowest \harmful hit rates, and models pretrained on \IG continue among the top \harmful label predictors. Notably, \ssl-\IG leads to the highest \harmful hit rates. For \nharmful predictions, we observe a general trend that results in larger hit rate differences across different groups, no matter the training paradigm and data used.

\textbf{Effect of data size and domain.} Figures~\ref{fig:cc_label_assoc}--\ref{fig:miap_label_assoc} (bottom) show how data domain and data size affect the training of \ssl models on \CC and \MIAP, respectively. The effect of data size when stratifying results based on age outlines the same trends as for the gender-skin tone stratification. Overall, it appears that increasing the data size from $1$M to $1$B \IG images, drastically increases the \nharmful predictions and also lowers the \harmful ones, especially on \CC. However, the \ssl-\ImNet results in comparable or lower \harmful prediction hit rates than the best \IG-based model, further emphasizing that the data content might be more critical than the scale of the data to mitigate potentially harmful associations.\looseness-1

\begin{figure}[t!]
\centering
\begin{subfigure}{.48\textwidth}
    \centering
    
     \includegraphics[width=1.\textwidth]{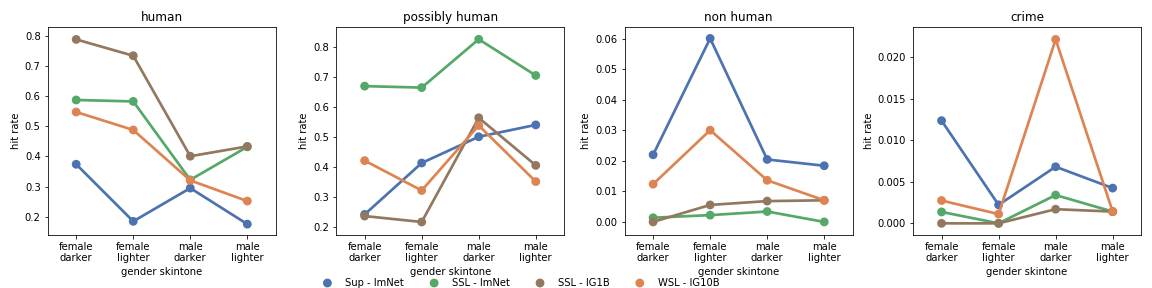} 
     
     \includegraphics[width=1.\textwidth]{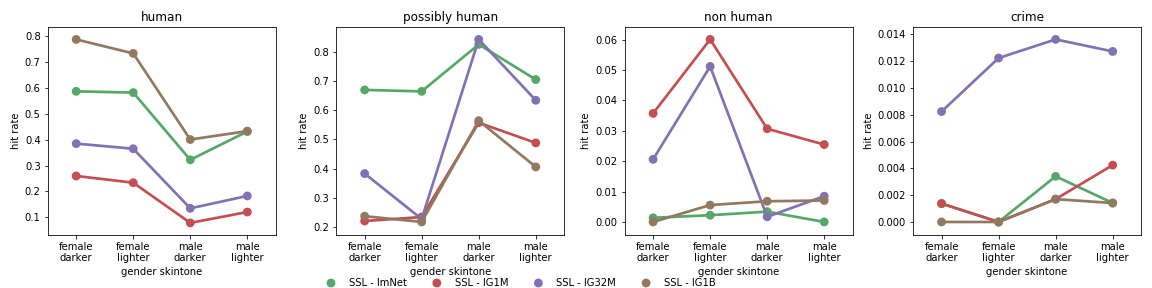}    

    \caption{Results based on gender and skin tone.}
    \label{fig:extended_gender_cc_label_assoc}
\end{subfigure}
\begin{subfigure}{.48\textwidth}
    \centering
     \includegraphics[width=1.\textwidth]{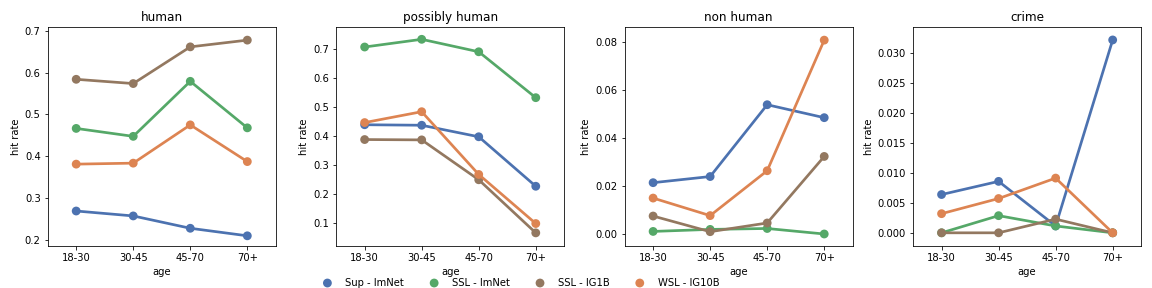}  
     
     \includegraphics[width=1.\textwidth]{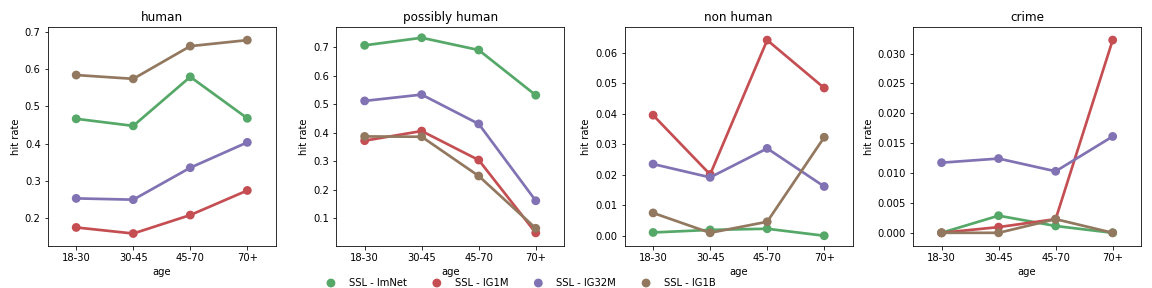}    

    \caption{Results based on age.}
    \label{fig:extended_age_cc_label_assoc}
\end{subfigure}
\caption{\CC extended label association results: (Top)  Effect of training paradigm; (Mid) Effect of data size and domain. For \human and \phuman, the higher the better. For \crime and \nhuman, the lower the better.\looseness-1 }
\label{fig:extended_cc_label_assoc}
\end{figure}

\begin{figure}[t!]
\centering
\begin{subfigure}{.45\textwidth}
    \centering
     \includegraphics[width=1.\textwidth]{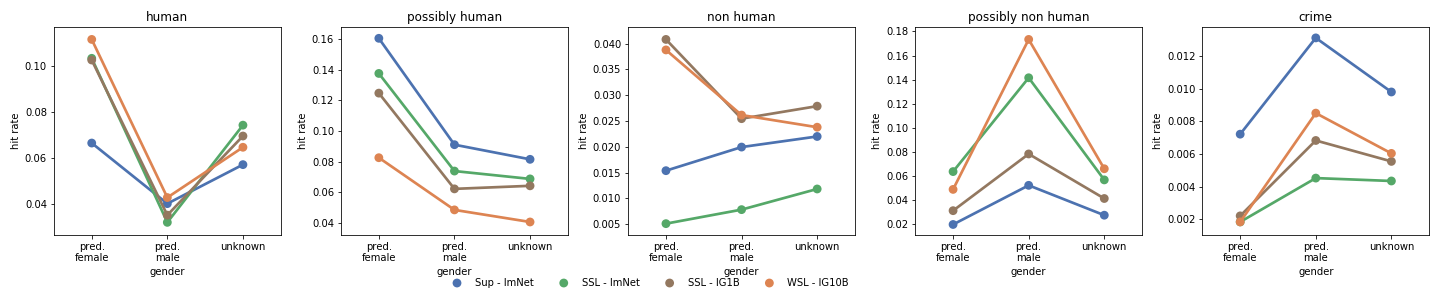}  
     
     \includegraphics[width=1.\textwidth]{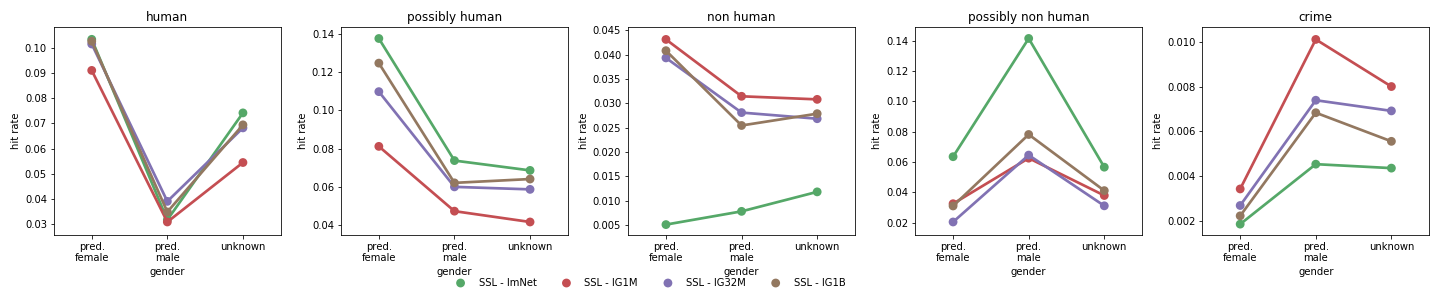}    

    \caption{Results based on perceived gender.}
    \label{fig:extended_gender_miap_label_assoc}
\end{subfigure}
\begin{subfigure}{.45\textwidth}
    \centering
    
     \includegraphics[width=1.\textwidth]{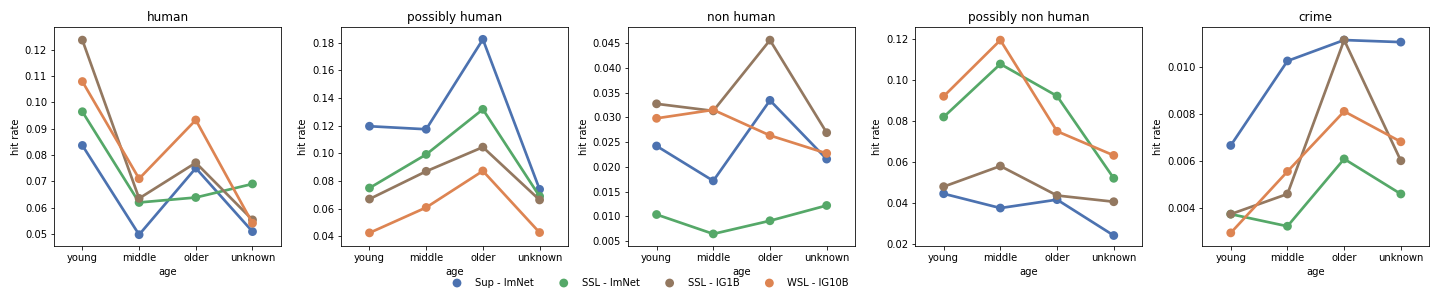} 
     
     \includegraphics[width=1.\textwidth]{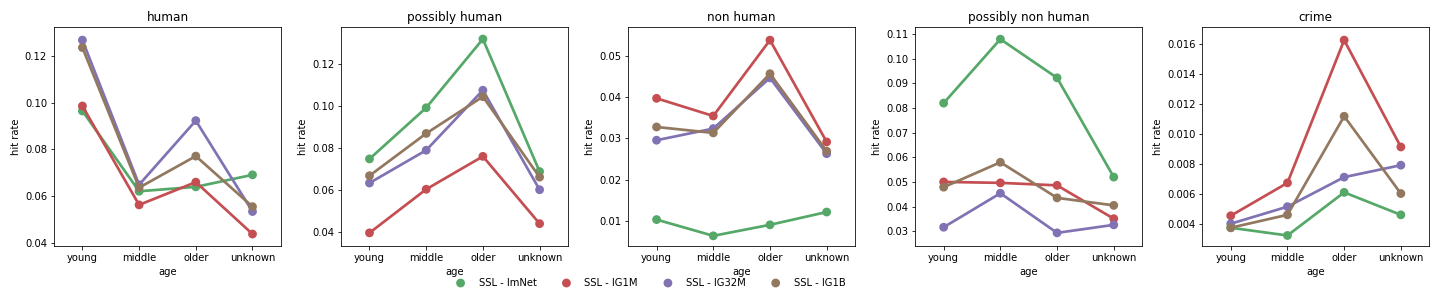}    

    \caption{Results based on perceived age.}
    \label{fig:extended_age_miap_label_assoc}
\end{subfigure}
\caption{\MIAP extended label association results: (Top)  Effect of training paradigm; (Mid) Effect of data size and domain; (Bottom) Confidence of models.For \human and \phuman, the higher the better. For \crime, \nhuman and \pnhuman, the lower the better.\looseness-1 }
\label{fig:extended_miap_label_assoc}
\end{figure}

For completeness, Figure~\ref{fig:extended_cc_label_assoc} presents label association results on \CC for all labels separately: \human, \nhuman, \phuman, and \crime. Results are shown both for gender and skin tone, as well as age, and consider the effect of training paradigm (top) and data size/domain (bottom). We note \ssl leads to both lower \crime and \nhuman hit rates. However, \ssl-\ImNet resulting in overall lower \harmful hit rates can be explained by its lower \nhuman prediction rates. \ssl-\IG results in lower \crime prediction rates, and both its \crime and \nhuman prediction rates can be decreased by appropriately increasing the dataset size (see IG1M vs IG1B results). Interestingly, \ssl-\ImNet consistently exhibits among the highest \phuman hit rates. Analogously, Figure~\ref{fig:extended_miap_label_assoc} presents extended results on \MIAP. In this case, results are reported for \human, \nhuman, \phuman, \pnhuman, and \crime, and consider both perceived gender and age. When decoupling the \harmful labels, we observe that the high \harmful prediction rates of \IG-pretrained models are driven by their frequent \nhuman predictions. However, when it comes to \crime, \supervised-\ImNet leads to the highest prediction rates. In \MIAP, \pnhuman predictions are more present in \ssl models, whereas \phuman predictions are more present in models pre-trained on \ImNet (no matter the paradigm).

\begin{figure}[t!]
\centering
\begin{subfigure}{0.45\textwidth}
    \centering
    
     \includegraphics[width=1.\textwidth]{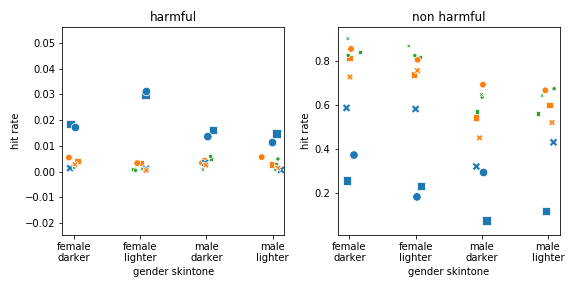} 
     
     \includegraphics[width=1.\textwidth]{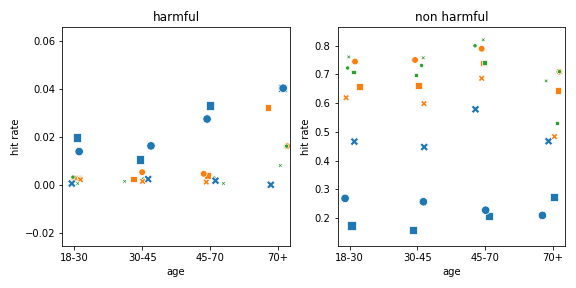}    
     
    \end{subfigure}
\begin{subfigure}{.45\textwidth}
    \centering
     \includegraphics[width=1.\textwidth]{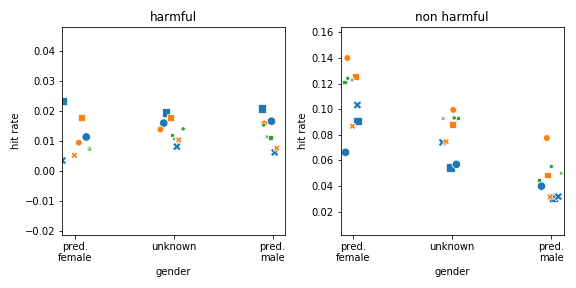}  
     
     \includegraphics[width=1.\textwidth]{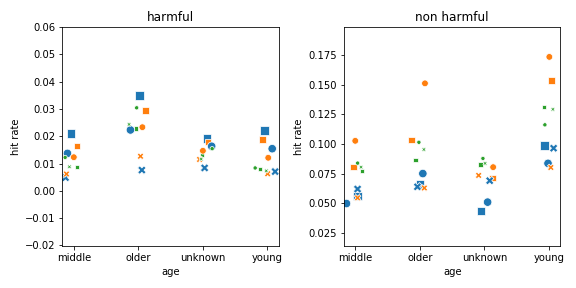}

\end{subfigure}
\begin{subfigure}{.9\textwidth}
    \includegraphics[width=1.\textwidth]{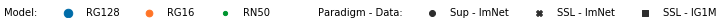}    
\end{subfigure}
\begin{subfigure}{0.45\textwidth}
    \centering
    \caption{\CC: (top) results based on gender and skin tone, (bottom): results based on age.}
    \label{fig:cc_label_assoc_capacity}
\end{subfigure}
\begin{subfigure}{.45\textwidth}
    \centering
    \caption{\MIAP: (top) results based on perceived gender, (bottom): results based on perceived age.}
    \label{fig:miap_label_assoc_capacity}
\end{subfigure}

\caption{Effect of model capacity on label association results.\looseness-1 }
\label{fig:label_assoc_capacity}
\end{figure}

\subsection{Effect of model capacity.}
\label{sec:label_assoc_model_capacity}
Finally, Figure~\ref{fig:label_assoc_capacity} presents the impact of model capacity on label association results on both \CC and \MIAP. When considering \supervised-\ImNet models, we observe that increasing model capacity tends to increase the prediction of \harmful labels across all groups considered in both \CC and \MIAP datasets. At the same time, the highest capacity model, \regnetbig, exhibits the lowest \nharmful hit rates. This suggests that better fitting the \ImNet data with a supervised objective may lead to undesirable outcomes. When considering \ssl-\ImNet models, increasing capacity maintains or decreases \harmful hit rates across all groups and for both datasets considered. It is worth noting that these \ssl models not only tend to exhibit among the lowest \harmful hit rates but also tend to be more stable in their predictions across different subgroups. However, higher capacity \ssl-\ImNet models often result in the lowest \nharmful hit rates. By contrast, when considering \ssl-\IG models, increasing model capacity often increases \harmful predictions and leads to slightly higher discrepancies across different groups. In this case, similar to \ssl-\ImNet, we observe that increasing model capacity often decreases or maintains the \nharmful predictions. However, as discussed in Section~\ref{sec:label_association}, these models can significantly benefit from additional \IG data to mitigate potentially \harmful label associations. It is worth noting that in the \MIAP, \nharmful prediction rates are overall significantly lower than in \CC...

\section{Effect of model capacity: geographical diversity}
\label{sec:geodiversity_model_capacity}

\begin{figure}[t!]
\centering
\includegraphics[width=\linewidth]{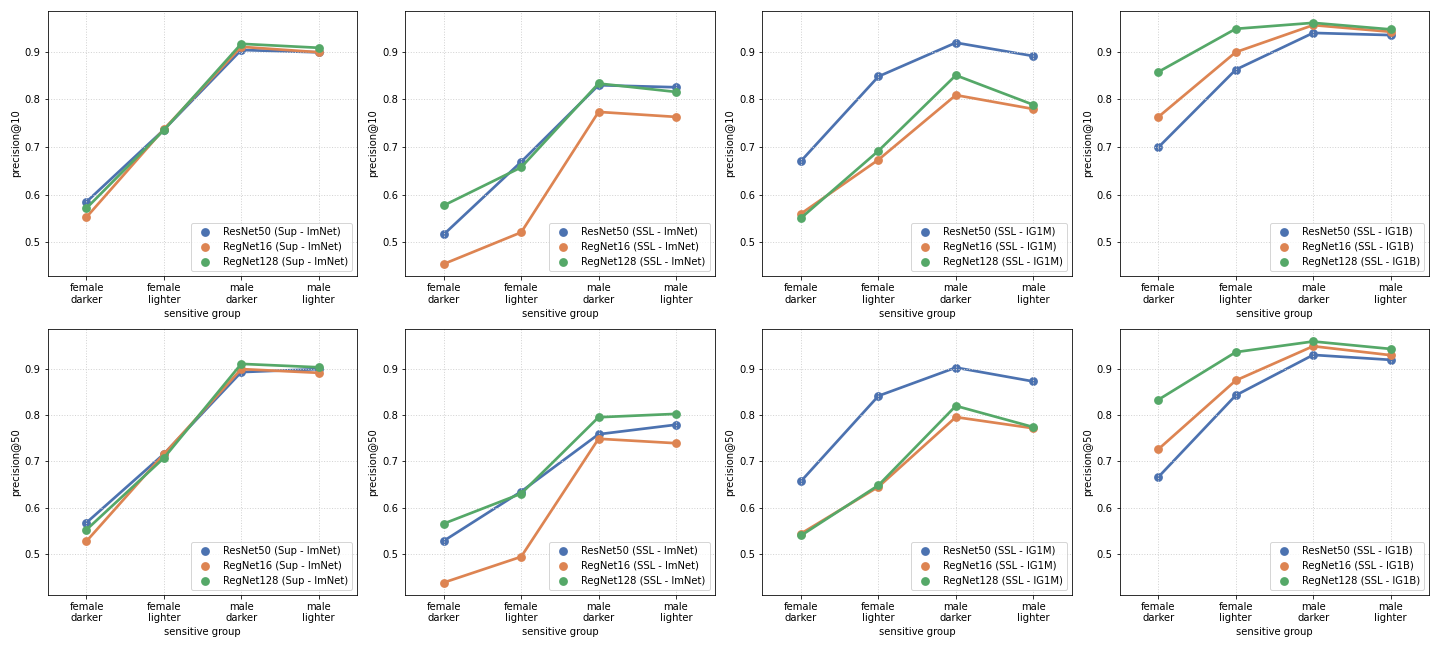}
\caption{Geodiversity: effect of model capacity depending on the training paradigm and data size (columns), broken down by income bucket (top row) and region (bottom row). \label{fig:geodiversity:model_capacity}}
\end{figure}

We provide in Fig.~\ref{fig:geodiversity:model_capacity} the results on the geographical fairness indicator depending on the model size, depending on the training paradigm (\supervised or \ssl) and depending on data domain and size (\ImNet, or \IG with $1m$, $1b$). 

We observe that model capacity is critical when pre-training in a supervised fashion on \ImNet (leftmost column): from \resnet to \regnetsmall there is an improvement of over $15\%$ (absolute) hit rate. \regnetsmall and \regnetbig achieve the same performance on \supervised training. Interestingly, model capacity does not seem to have a large impact for \ssl on $1m$ images, either on \ImNet or \IG (middle left and middle right columns), and the performances dominate that of \supervised pre-training. The effect of model capacity becomes more visible when training on $1b$ images (rightmost column), where differences between \regnetbig and \resnet are significant across all sensitive groups. 

The conclusion is that the results coincide with those of other indicators when training at very large scale: larger models tend to fare better. Apart from that, the effect of model size on supervised training is interesting, yet seems specific to that indicator. 

\section{Effect of model capacity: similarity search}
\label{sec:simsearch_model_capacity}

\begin{figure}[t!]
\centering
\includegraphics[width=\linewidth]{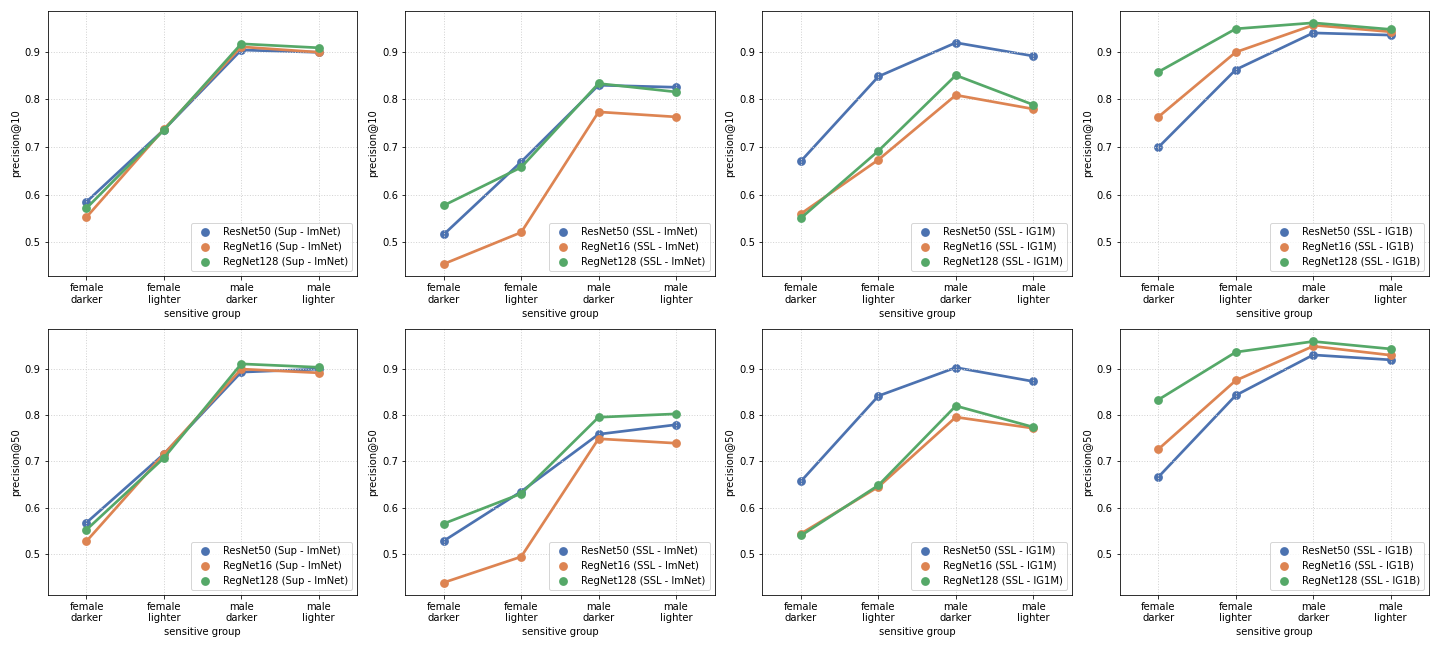}
\caption{Similarity search indicator: precision@10 (top row) and precision@50 (bottom row) depending on model capacity, for different training paradigms, data domain and sizes (columns). \label{fig:simsearch:model_capacity}}
\end{figure}

We present in this section the results obtained on the similarity search indicator by varying the model capacity. Fig.~\ref{fig:simsearch:model_capacity} presents the results in terms of precision@10 and precision@50 for \resnet, \regnetsmall and \regnetbig for \supervised training on \ImNet (left most column), \ssl on \ImNet (middle-left column), \ssl on \IG with $1m$ examples (middle-right) and \ssl on \IG with $1b$ examples. 

As for all other indicators, increasing model capacity improves precision on \ssl training on $1b$ examples, with \regnetbig achieving the best precision values (and about $15\%$ absolute improvement on \femaledarker, the sensitive group where the precision is the lowest). 

For other training paradigms/data sizes however, increasing model capacity does not help much. On \supervised pre-training, we see no effect of increasing the model capacity, while when using \ssl on $1m$ examples (\ImNet or \IG) \resnet tends to perform the best (on par with \regnetbig on \ImNet, by far the best on \IG with $1m$ examples). In conclusion, it seems that on this indicator, large-scale datasets are necessary for \ssl with large models to shine. Yet, as noticed in Sec.~\ref{sec:simsearch}, compared to supervised pre-training, the improvement  of \ssl on \IG with $1b$ examples is substantial (comparing the leftmost and rightmost columns), with nearly $30\%$ absolute improvement in precision@50 on \femaledarker between supervised pre-training on \ImNet and \regnetbig trained with \ssl.

\end{document}